\documentclass[10pt,twocolumn,letterpaper]{article}

\usepackage[pagenumbers]{iccv} 

\usepackage{multirow}

\usepackage{colortbl}
\usepackage{lipsum}
\usepackage{bm}
\usepackage{tikz}
\usepackage{pgfplots}
\pgfplotsset{compat=1.18} 
\usepackage{pifont}
\usepackage{array}
\usepackage{graphicx}
\usepackage{rotating}
\usepackage{verbatim}
\usepackage{tikz}
\usepackage[pagebackref,breaklinks,colorlinks,allcolors=iccvblue]{hyperref}
\usepackage{orcidlink}
\usepgfplotslibrary{groupplots}

\definecolor{firstcolor}{HTML}{BDE6CD}
\definecolor{secondcolor}{HTML}{E2EEBC}
\definecolor{thirdcolor}{HTML}{FFF8C5}

\newcommand{\fst}[1]{\cellcolor{firstcolor}\bfseries #1}
\newcommand{\snd}[1]{\cellcolor{secondcolor} #1}
\newcommand{\trd}[1]{\cellcolor{thirdcolor}#1}
\newcommand{\FAIL}[0]{{\color{red}\ding{55}}\ }

\newcommand{\cmark}{{\color{PineGreen}\checkmark}}
\newcommand{\xmark}{{\color{red}\ding{55}}\ }

\newcommand{\method}[0]{ToF-Splatting\xspace}

\newcommand{\da}[0]{$\downarrow$}
\newcommand{\ua}[0]{$\uparrow$}

\newcommand{\boldparagraph}[1]{\vspace{0pt}\noindent{\bf #1}}

\newcommand{\smatt}[1]{\color{violet} #1 \color{black}}
\newcommand{\andrea}[1]{\color{teal} #1 \color{black}}
\newcommand{\martin}[1]{\color{blue} #1 \color{black}}
\newcommand{\valerio}[1]{\color{green} #1 \color{black}}
\newcommand{\matteo}[1]{\color{orange} #1 \color{black}}

\definecolor{iccvblue}{rgb}{0.21,0.49,0.74}

\newcommand\blfootnote[1]{%
  \begingroup
  \renewcommand\thefootnote{}\footnote{#1}%
  \addtocounter{footnote}{-1}%
  \endgroup
}

\def\paperID{12384} 
\def\confName{ICCV}
\def\confYear{2025}

\makeatletter
\renewcommand{\and}{\hspace{2pt}}  
\makeatother

\title{ToF-Splatting: Dense SLAM using Sparse Time-of-Flight Depth\\ and Multi-Frame Integration}
\author{
Andrea Conti 
$^{1, 2}$\footnote{work started while visiting University of Amsterdam} {}
\and Matteo Poggi 
$^{1}$ 
{}
\and Valerio Cambareri 
$^{2}$ {}
\\
\and Martin R. Oswald 
$^{3}$ {}
\and Stefano Mattoccia 
$^{1}$ {}
\\
{\small $^1$University of Bologna, Italy \ $^2$Sony DepthSensing Solutions, Belgium \ $^3$University of Amsterdam, Netherlands}
}

\begin{document}

\twocolumn[{%
  \renewcommand\twocolumn[1][]{#1}%
  \maketitle
  \centering
  \captionsetup{type=figure} 
  \includegraphics[width=0.9\linewidth]{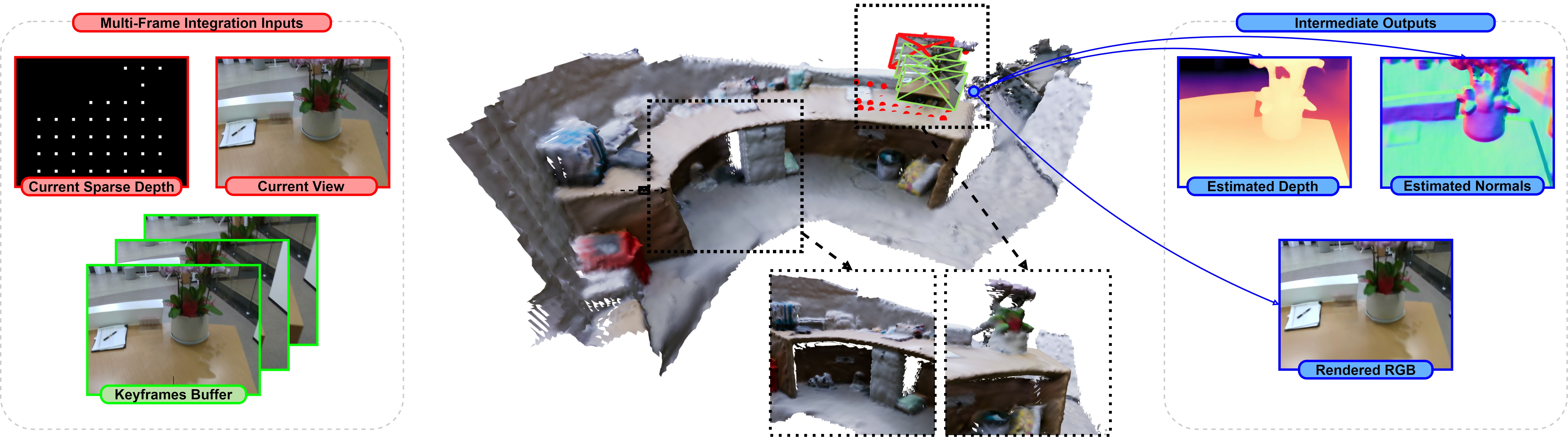}\\
  \captionof{figure}{\textbf{Overview of our ToF-Splatting method.} Our method combines sparse ToF depth, multi-view geometry from a buffer of keyframes, and monocular cues (left) to perform into a unique end-to-end dense SLAM framework enabled by a Gaussian Splatting.}
  \label{fig:teaser}
  \vspace{0.354cm}
}]

\begin{abstract}

Time-of-Flight (ToF) sensors provide efficient active depth sensing at relatively low power budgets; among such designs, only very sparse measurements from low-resolution sensors are considered to meet the increasingly limited power constraints of mobile and AR/VR devices. However, such extreme sparsity levels limit the seamless usage of ToF depth in SLAM. In this work, we propose \method, the first 3D Gaussian Splatting-based SLAM pipeline tailored for using effectively very sparse ToF input data. Our approach improves upon the state of the art by introducing a multi-frame integration module, which produces dense depth maps by merging cues from extremely sparse ToF depth, monocular color, and multi-view geometry. Extensive experiments on both real and synthetic sparse ToF datasets demonstrate the advantages of our approach, as it achieves state-of-the-art tracking and mapping performances on reference datasets. 

\end{abstract}    
\section{Introduction}
\label{sec:intro}
\blfootnote{$^*$ work started while visiting University of Amsterdam.}Simultaneous Localization and Mapping (SLAM)~\cite{whelanElasticFusionRealtimeDense2016,daiBundleFusionRealtimeGlobally2017,zilong_slam,consistent_depth} entails the joint estimation of the current camera pose against a 3D scene representation (\emph{i.e.}, the map) and the update of this representation with data from the current camera view. It is a fundamental task with applications in AR/VR, robotics, and autonomous navigation~\cite{ptam,rk_slam}. 
In recent years, SLAM witnessed a revolution \cite{tosi2024nerfs3dgaussiansplatting} following the success of Neural Radiance Fields (NeRF) \cite{mildenhallNerfRepresentingScenes2020} and 3D Gaussian Splatting (3DGS) \cite{Kerbl20233DGS} as scene representations from the adjacent field of novel view synthesis. 

In this paper we focus on SLAM pipelines that leverage RGB-D video inputs, \emph{i.e.}, where color images and dense depth maps are collected by synchronized, accurate color and depth sensors. Beyond global scale, depth sensors provide accurate local geometry that grants better tracking and mapping quality. A common choice for active depth sensing are Time-of-Flight (ToF) cameras. Despite the increased accuracy provided by reliable depth measurements, the power consumption and cost of such sensors constrain their adoption to high-end industrial or automotive applications. A recent trend to favor ToF sensor integration in handheld consumer devices is to reduce the specifications of ToF sensors to provide low-resolution but more reliable depth measurements. Among other examples, LiDAR (an instance of ToF) appears in consumer mobile phones~\cite{luetzenburg_evaluation_2021} and is known to feature up to at most $576$ dots (interpolated to $256~\times~192$ resolution via depth completion). Other recent ToF sensors provide only up to $64$ depth points but consume a reported $200~mW$~\cite{tof-slam}. In such settings a SLAM pipeline can expect to receive, \emph{e.g.}, $8~\times~8$ very noisy, very sparse depth measurements per frame. These specifications prevent their straightforward integration in SLAM systems, as low-resolution noisy depth is known to harm the accuracy of dense SLAM systems built with NeRF and not originally designed to deal with this scenario~\cite{whelanElasticFusionRealtimeDense2016,zhuNICESLAMNeuralImplicita,sucar2021imap}. This is even more evident with the more recent 3DGS-based SLAM systems \cite{monogs}, where accurate and dense depth is paramount to achieve satisfactory results.

To overcome these limitations, a straightforward strategy~\cite{tofslam} would consist in recovering dense and accurate depth maps using a single-image depth completion framework~\cite{qiao2024rgb}. However, models trained for this task are known to generalize poorly across domains and across input depth noise levels, potentially affecting the downstream SLAM pipeline accuracy and yielding both drifted camera trajectory and inaccurate mapping. Therefore, we explore 3DGS as an explicit representation that allows for injecting domain-specific knowledge and that is easily extended to, \eg, dynamic scenes~\cite{Luiten2023Dynamic3G} 
and accurate large mesh reconstruction~\cite{guedon2023sugar, chen2024vcrgaus}. 
On the one hand, we argue that depth completion alone is insufficient to allow sparse ToF depth usage in SLAM systems. On the other, we acknowledge that even sparse and noisy depth measurements can bootstrap SLAM if properly assisted with geometry. Accordingly, inspired by frame-to-frame SLAM systems relying on pre-trained trackers for localization, we believe that \textit{multi-frame integration} fusing the noisy depth measurements with multi-view geometry across a small set of local frames could supply the dense SLAM system with the supervision necessary to fill this gap. 

In this paper, we propose \method, a novel 3DGS-based SLAM system enabling dense reconstruction from sparse ToF sensors. \method harmoniously alternates tracking and mapping, with each one benefiting from the other. The former is carried out through backpropagation during the optimization of the 3DGS model on color images; the latter exploits a pre-trained network~\cite{depth-on-demand} as a multi-frame integration module, combining sparse depth measurements with multi-view geometry according to a buffer of keyframes. For each of the keyframes, we use the camera pose estimated through tracking and providing dense depth maps to improve the supervision given to the 3DGS model itself. Figure~\ref{fig:teaser} shows an overview of the setting in which \method{} operates, as well as the quality of the 3D reconstructions it yields.

\method is tested on the real-world ZJUL5 dataset~\cite{tof-slam}, where we measure tracking and mapping quality and find the superiority of our framework with respect to existing solutions built over sparse ToF sensors~\cite{tofslam}. Moreover, standard benchmarks such as Replica~\cite{replica19arxiv} are also provided and confirm the effectiveness of our mapping strategy. Our main \textbf{contributions} are as follows:
\begin{itemize}
    \item We propose \method, the first 3DGS-based SLAM system suited for sparse ToF sensors.
    \item We introduce a novel multi-frame integration method combining sparse depth, multi-view geometry, and monocular cues for robust noise and outlier handling.
    \item We establish a new state-of-the-art for dense SLAM systems built on top of sparse ToF depth sensors.
\end{itemize}
%


\begin{figure*}[!th]
    \centering
    \includegraphics[trim=0cm 11cm 5cm 0cm, clip, width=0.92\linewidth]{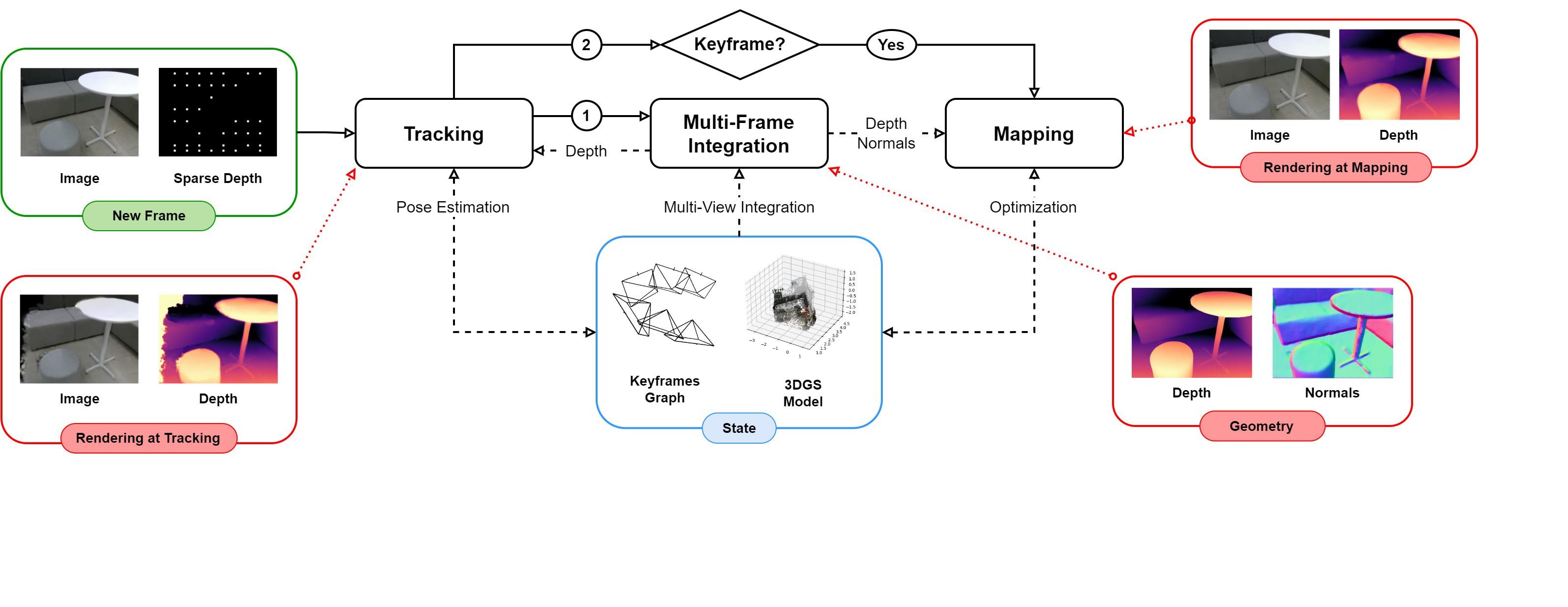}\vspace{-0.2cm}
    \caption{\textbf{\method{} Pipeline.} Our method involves three main modules: a Tracking frontend estimating camera poses, a Multi-Frame Integration module that predicts dense depth maps from sparse ToF measurements and multi-view geometry, and a Mapping backend modeling the 3D scene representation via 3D Gaussian Splatting.}\vspace{-0.3cm}
    \label{fig:overview}
\end{figure*}

\section{Related Work}
\label{sec:related}

We briefly review the literature relevant to our work, referring to \cite{tosi2024nerfs3dgaussiansplatting} for a detailed overview of the latest advances.

\boldparagraph{Traditional SLAM.} Traditional visual SLAM pipelines, or the more recent ones introducing deep modules \cite{tateno2017cnn, li2020structure, teed2021droid}, usually perform tracking in frame-to-frame fashion, based on the visual features extracted from past frames either using handcrafted feature extractors \cite{mur2017orb} or deep networks \cite{tateno2017cnn,teed2021droid}. These systems are usually defined by four main steps: tracking, mapping, global bundle adjustment, and loop closure. 
Among the different representations used by traditional SLAM frameworks, depth points \cite{mur2017orb, campos2021orb}, 
surfels \cite{schops2019bad}, and volumetric representations \cite{daiBundleFusionRealtimeGlobally2017} allows for achieving globally consistent 3D reconstruction.

\boldparagraph{NeRF-based SLAM.} The advent of NeRF in the field of novel view synthesis also reached adjacent research areas, as well as SLAM a few years later. NeRF-based SLAM systems estimate camera poses and reconstruct dense meshes by modeling scenes via MLPs. iMAP \cite{sucar2021imap} and NICE-SLAM \cite{zhu2022nice} are the first SLAM frameworks following this approach, with the latter introducing dense feature grids supporting the MLPs. Point-SLAM \cite{sandstrom2023point} and Loopy-SLAM \cite{liso2024loopy} switched feature grids with neural point embeddings, allowing for higher flexibility and correcting or adjusting local maps after global optimization, yet with slow processing prohibiting deployment in robotics. 
ESLAM \cite{johari2023eslam} and Co-SLAM \cite{wang2023co} deploy tri-planes and hash grid embeddings, improving both processing speed and reconstruction accuracy, with PLGSLAM \cite{deng2024plgslam} further improving the representation capacity in large indoor scenes. 
GO-SLAM \cite{zhang2023go} and KN-SLAM \cite{wu2024kn} implement frame-to-frame systems thanks to external trackers, yielding a good trade-off between tracking/mapping accuracy and frame rate.

\boldparagraph{3DGS-based SLAM.} More recent advances in novel view synthesis brought to the development of new SLAM systems using 3D Gaussian Splatting \cite{Kerbl20233DGS} to represent the mapped scene. 
Several concurrent frameworks emerged built over 3DGS, such as MonoGS \cite{monogs}, GaussianSLAM \cite{yugay2023gaussianslam}, SplaTAM \cite{keetha2024splatam}, and GS-SLAM \cite{yan2024gs}. 
Different from NeRF-based SLAM systems, these frameworks allow for higher interpretability and higher rendering speed.

\boldparagraph{Sparse ToF-based SLAM.} Only lately, the use of more compact and energy-efficient depth sensors has gained attention in the SLAM research community. ToF-SLAM \cite{tofslam} represents the only attempt to exploit ``lightweight'', sparse depth measurements within a dense SLAM system, aided by a depth completion model \cite{deltar}. However, depth completion alone fails to deal with noise in the sparse input depth, a shortcoming we aim to specifically overcome within our \method framework, the first one based on 3DGS and suited for sparse ToF sensors. 
\section{Pipeline Architecture}
In this section, we present the core components of our framework. 
\method is a frame-to-model SLAM system conceptually divided into three modules: $(i)$ a mapping backend, $(ii)$ a tracking frontend, and $(iii)$ a peculiar multi-frame integration module yielding dense depth maps based on a subset of the keyframe graph, the current RGB frame, and the current sparse ToF depth map. 

Figure \ref{fig:overview} provides an overview of our pipeline. Our framework is characterized by three heterogeneous modules that are tightly related and influence each other. The tracking module stands as the most influencing element in the pipeline, as it selects the keyframes used by both the mapping and multi-frame integration parts, moreover, it computes the poses used for multi-frame depth estimation. Nonetheless, it relies on the quality of the mapping step to perform correct ego-motion estimation and exploits the depth cues to properly behave when photometric information is lacking or ambiguous. 

The depth perception part is mainly sustained by the tracking step and highly influences the mapping part since it is used to seed new Gaussians and as supervision. Finally, the mapping phase not only creates a unique smooth representation but also defines the rendered opacity that is used for keyframe selection. These connections lead to a smoothly integrated framework that not only exploits a multi-frame backbone but more importantly, defines an effective way to use such information in the SLAM scenario.


\subsection{Background: 3D Gaussian Splatting}
\label{sec:gaussian-splatting}

3D Gaussian Splatting \cite{Kerbl20233DGS} (3DGS) represents one of the latest advances in novel view synthesis, fitting a dynamically defined set of multivariate Gaussian distributions over the scene to enable image rendering from arbitrary viewpoints. Each Gaussian $G(\mathbf{x})$ is parametrized by its mean $\bm\mu \in \mathbb{R}^3$ and covariance matrix $\bm\Sigma \in \mathbb{R}^{3\times3}$ , reading
\begin{equation}
    G(\mathbf{x}) = \exp\left(\ {-\frac{1}{2}(\mathbf{x}-\bm\mu)^\top\bm\Sigma^{-1}(\mathbf{x}-\bm\mu)}\ \right)
    \label{eq:multivariate-gaussian}
\end{equation}
Moreover, opacity $o \in \mathbb{R}$ and RGB color $\mathbf{c} \in \mathbb{R}^3$ are assigned to each Gaussian.

Given a set of images from different viewpoints all around the scene, the Gaussian parameters are optimized using the splatting technique \cite{Wang2019DifferentiableSS}. With such a method, 3D Gaussians are projected through rasterization over the 2D image plane, containing a 2D Gaussian distribution  
\begin{equation}
    \bm\mu^{2D} = \pi(\mathbf{T}_{wc}\bar{\bm\mu}) \qquad \bm\Sigma^{2D} = \mathbf{J}\mathbf{R}_{wc}\bm\Sigma\mathbf{R}_{wc}^\top\mathbf{J}^\top
    \label{eq:splatting}
\end{equation}
where $\bar{\bm\mu}$ is $\bm\mu$ in homogeneous coordinates, $\mathbf{T}_{wc} \in \mathbb{R}^{4\times4},\ \mathbf{R}_{wc} \in \mathbb{R}^{3\times3}$ are respectively the world-to-camera transformation and its rotational component, $\pi$ is the projection matrix and $\mathbf{J} \in \mathbb{R}^{2\times3}$ is the Jacobian of the projective transformation \cite{Zwicker2001SurfaceS}. The color $C$ of a pixel is then given by the combination of the 
$M$ weighted Gaussians, \emph{i.e.}, 
\begin{equation}
    \mathbf{C} = \sum_{i\in M} \mathbf{c}_i\alpha_i\prod_{j=1}^{i-1}(1 - \alpha_j)
\end{equation}
with $\alpha_i = o_iG(\mathbf{x}_i)$. Since $\bm\Sigma$ is positive semi-definite, it is parametrized with a diagonal scaling matrix $\mathbf S$ and a rotation matrix $\mathbf R$, so that $\bm\Sigma = \mathbf{R}\mathbf{S}\mathbf{S}^\top\mathbf{R}^\top$.  $\mathbf R$ is internally represented with quaternions.

\subsection{Multi-Frame Integration}
\label{sec:multi-view-depth-estimation}

Dense depth cues are pivotal for proper initialization of the Gaussians position $\bm\mu \in \mathbb{R}^3$, as well as to provide supervision to both tracking and mapping threads -- as already known in the literature \cite{monogs,nicer-slam}. 
However, 
sparse ToF sensors alone are insufficient for this purpose, because of the very sparse and noisy depth measurements they provide (\eg, only 64 points for a sensor such as that used in \cite{tof-slam}).
To overcome this limitation, ToF-SLAM \cite{tof-slam} exploits 
an auxiliary completion model, DELTAR \cite{deltar}, to recover dense depth maps out of ToF measurements. However, depth completion is highly dependent on the quality of the sparse depth in input, rapidly degrading the accuracy of the densified depth maps in the presence of noise. 
%

In this paper, we follow a different path: inspired by frame-to-frame SLAM systems deploying an external tracker, we believe that multi-frame integration can compensate for the noise in the ToF measurements by exploiting both single-view and multi-view geometry cues on a set of local frames, given the camera poses estimated by the tracking frontend. We choose to extend the Depth on Demand (DoD) framework~\cite{depth-on-demand} for this purpose.


\boldparagraph{Multi-Frame Integration.} DoD~\cite{depth-on-demand} 
integrates sparse depth and two-view stereo cues for dense depth prediction. Specifically -- given a frame $\mathbf{F}^k$ -- it processes the associated RGB image $I_k$ (the \emph{target view}), another RGB view $I_j$ from a different frame $\mathbf{F}^j$ (a \emph{source view}), the relative pose $P^{jk}$ between the two, and the sparse depth points $H^k$ to predict a dense metric depth map $D^k$. Differently from monocular depth prediction, DoD is aware of the scene scale either from $H^k$ or $P^{jk}$, simplifying the problem of integrating scale-inconsistent geometries in the SLAM frontend. 
However, DoD is limited to two-view processing: this may lead to suboptimal results in the SLAM scenario, where an abundant amount of keyframes largely overlap and cover the same portion of the scene. 
Thus, we extend~\cite{depth-on-demand} to multi-view processing. This is achieved by observing that the predicted depth map $D^k$ depends on the source view $I_j$ through its relative pose $P^{jk}$ only. DoD iteratively updates $D^k$ for multiple iterations, exploiting this intrinsic property we can integrate multiple views $(I_j, P^{jk})$ using a different one at each iteration~\cite{ramdepth}. 
This approach allows smooth integration of multi-view cues avoiding order dependency. To select the source views $I_j$, at each prediction we order the keyframes selected by the SLAM pipeline according to their relative pose similarity and select the first $N$ frames with baseline distance close to $b = 15 \text{cm}$ to ensure enough parallax. The model is trained from scratch on ScanNetv2~\cite{dai2017scannet} and following the protocol of~\cite{depth-on-demand}.

\boldparagraph{Monocular Cues Integration.} 
As shown in \cite{depth-on-demand}, DoD heavily relies on both geometry and depth measurements, thus struggling at generalizing in challenging scenarios where the two lack.
This may happen in cases where the tight field of view, noisy sparse points, and textureless surfaces hamper multi-view matching. 
To mitigate such limitations, we integrate explicit monocular cues into DoD by feeding its monocular encoder with $I_k$ and the normalized depth map $\tilde{D}^k \in [0, 1]^{H \times W}$ obtained through the single-image depth estimator Depth Anything~v2~\cite{depth-anything-v2}. This approach allows the injection of a robust bias, avoiding issues related to monocular estimation, \eg, slanted surfaces or noisy scale estimates, and delegating to DoD their handling. These cues are included while retraining on ScanNetv2 \cite{dai2017scannet}.

\boldparagraph{Outlier Handling.} Furthermore, the noisy depth measurements in input to DoD can severely affect its accuracy.
Therefore, we explicitly focus on dealing with outliers in the input sparse depth, that may be perceived by ToF devices over specific surfaces. Such errors are particularly evident in cheap sensors and may largely prejudice the information supplied by the few points they measure. Purposely, we deploy a simple yet effective method to identify such outliers by exploiting multi-view geometry again. Whenever depth needs to be predicted for a new frame, DoD predicts a depth map $\hat{D}^k$ without processing the ToF measurements. Then, the $\ell_1$ error between the latter and the depth measured by the ToF sensor $H^k$ is computed
%
%
and finally sparse measures 
with error higher than a given quantile $q$ are discarded, and depth is predicted again by DoD by also processing the filtered sparse depth $\tilde{H}^k$ this time. 
This way, we exploit multi-view cues to extract a clean depth prior to filtering out inconsistent sparse depth measures, and then we integrate the remaining points to improve the original prediction. Even without processing the sparse depth, DoD still predicts metric depth, as the metric scale is enforced by the relative camera poses.

\subsection{Tracking Frontend}
\label{sec:tracking}

The tracking part of \method{} estimates the ego-motion of the camera for each new frame $\mathbf{F}^k$ and builds a keyframe graph $\mathbf{G} = \{ \mathbf{KF}^k \} \subset \{ \mathbf{F}^k \}$ containing a set of meaningful frames for mapping and multi-frame depth estimation.

\boldparagraph{Pose Estimation.} We initialize each new frame $\mathbf{F}^k$ of our frame-to-model tracker with the pose of the previous frame $\mathbf{F}^{k-1}$ and then minimizing the tracking loss with respect to the relative pose between $\mathbf{F}^k$ and $\mathbf{F}^{k-1}$, following~\cite{monogs}. The tracking loss consists of two terms, $L_{\text{rgb}}$ and $L_{\text{depth}}$, respectively accounting for photometric and geometric errors  
\begin{align}
    L_{\text{track}} &= \lambda_{\text{track}} L_{\text{rgb}} + (1 - \lambda_{\text{track}}) L_{\text{depth}} \\
    \label{eq:trackingloss}
    L_{\text{rgb}} &= \Vert I_{k} - \hat{I}(\textbf{F}^k) \Vert_1 \quad  \ L_{\text{depth}} = \Vert D^k - \hat{D}(\textbf{F}^k) \Vert_1
\end{align}
where $I_k$ is the RGB image from $\textbf{F}^k$, $\hat{I}(\cdot)$ and $\hat{D}(\cdot)$ denote the rendered RGB and Depth and, $D^k$ is the depth prediction produced by the multi-frame integration module.

Unlike other methods, the tracking process happens in two stages. Initially, for a fixed number of steps $\eta_{\text{rgb}}$ only the photometric loss $L_{\text{rgb}}^T$ is optimized. Then, such a pose is used for multi-frame integration, as described in Sec.~\ref{sec:multi-view-depth-estimation}, to generate $D^k$. Finally, both losses in Eq.~\eqref{eq:trackingloss} are used for $\eta_{\text{rgbd}}^T$ further steps.


\boldparagraph{Keyframe Selection Policy.} We deploy a simple yet effective keyframe selection policy that considers i) novel content to be mapped and ii) instabilities in the already mapped areas. Unlike other methods using complex approaches -- \eg, frustum intersection, point clouds and pose analysis \cite{Yugay2023GaussianSLAMPD, monogs} -- we observe that the rendered opacity for a novel frame contains low opacity values where either Gaussians are not present or unstable. The first case highlights the presence of novel areas of the environment to be mapped, the second happens due to the pruning procedure involved in the mapping step (see Sec.~\ref{sec:mapping}). We set a frame $\mathbf{F}^k$ as a keyframe $\mathbf{KF}^k$ if its novelty factor $\nu^k$ is higher than a threshold $\nu^{th}$, where we define $\nu^k$ as
\begin{align}
    \nu^k &= \frac{\sum_{i \in \mathbf{KF}^k} \mathbf{M}(\mathbf{KF}^k_i)}{\sum_{i \in \mathbf{KF}^k} \big(1-\mathbf{M}(\mathbf{KF}^k_i)\big)} \\
    \label{eq:novelty-factor}
    \mathbf{M}(\mathbf{KF}^k_i) &= \begin{cases}
        \ 1 \quad \text{if} \quad \hat{O}(\mathbf{KF}^k_i) \ < \ \sigma \\
        \ 0 \quad \text{otherwise}\\ 
    \end{cases} 
\end{align}
with $\hat{O}(\mathbf{KF}^k_i)$ being the rendered opacity at pixel $i$, and $\mathbf{M}(\mathbf{KF}^k_i)$ a binary uncertainty map.

This approach allows for skipping several frames when the camera moves slowly or focusing deeply on areas where the \method{} struggles the most, by mapping it consecutively. Nonetheless, we enforce mapping after skipping a certain number of frames 
to avoid mapping too few frames.

\subsection{Mapping Backend}
\label{sec:mapping}

Whenever a new keyframe $\mathbf{KF}^k$ is identified, mapping is triggered. This phase seeks to embed the new frame in the global 3DGS model, involving the following two steps.

\boldparagraph{Initialization.} In this step, new Gaussians are seeded in the 3DGS model. To limit the size of the model and reduce outliers, only areas where the rendered uncertainty $\mathbf{M}(\mathbf{KF}^k)$ is high are seeded with new points, with the new Gaussians also being randomly downsampled by a constant factor. Each Gaussian is initialized using depth $D^k$ predicted in the multi-frame integration step, with color from the RGB frame $I_k$, and scale initialized with a constant value. 
Unlike other 3DGS pipelines~\cite{Yugay2023GaussianSLAMPD, monogs}, we prove this approach effective and particularly efficient, replacing any complex 3D heuristics to a simple use of rendered opacity $O(\cdot)$, already a side-product of the rasterization process.

\boldparagraph{Optimization.} Firstly, \method{} smoothly integrates the new information in the global model and secondly tunes the existing representation. This is achieved by optimizing the new keyframe $\mathbf{KF}^k$ and a random subset of 5 keyframes among the last $N$ keyframes $\{\mathbf{KF}^j\ \ | \ j \ge \max(k - N, 0)\}$. These are optimized for a fixed number of steps $\eta_{\text{rgbd}}^M$. Moreover, we enforce a sampling rate of 60\% for $\mathbf{KF}^k$, with the remaining 40\% for the remaining $\mathbf{KF}^j$ to avoid forgetting. 
During this step, gradients are also back-propagated to the camera poses of the sampled keyframes, thus acting as a global bundle adjustment.
Finally, we prune Gaussians with opacities lower than $0.5$ every $\eta_\text{rgbd}^M / 2$ steps. 
In addition to the terms in Eq.~\eqref{eq:trackingloss}, we optimize structural similarity as in Eq.~\eqref{eq:mappingloss} and, following \cite{chen2024vcrgaus}, the normals as in Eq.~\eqref{eq:normalsloss}, \emph{i.e.}, 
%
\begin{align}
    L_\text{dssim} & = \frac{1 - \text{SSIM}\big(I_k, \hat{I}(\textbf{KF}^k)\big)}{2}
    \label{eq:mappingloss} \\
    L_\text{normals} & = \Vert N^k - \hat{N} \Vert_1 + \overline{\textbf{1} - \langle N^k, \hat{N} \rangle} \label{eq:normalsloss}
\end{align}
with $\hat{N} := \bar\nabla \hat{D}(\textbf{KF}^k),\  N^k := \bar\nabla D^k$ denoting the normals estimated via 3D gradients $\bar\nabla$ for the rendered and multi-frame predicted depth respectively, $\langle \cdot,\cdot\rangle$ their inner product, and $\overline{\enskip\cdot\enskip}$ the average. $L_\text{normals}$ is masked on edges to preserve sharpness, as identified by running a Sobel filter. 

Finally, we introduce a regularization loss $L_\text{iso}$ to penalize elongated Gaussians (\emph{i.e.}, promote their isotropy) that usually leads to rendering artifacts as already observed in~\cite{monogs}. Eq.~\eqref{eq:mappingreg} defines $L_\text{iso}$, where $\text{diag}(\cdot)$ extracts the diagonal values from its argument, and $\overline{\text{diag}(\textbf{S}_j)}$ is the average of the resulting vector, \emph{i.e.}, of the elements of the $j$-th diagonal scale matrix $\textbf{S}_j$ as defined in Sec.~\ref{sec:gaussian-splatting}. Thus,
\begin{equation}
    L_\text{iso} = \frac{1}{|\mathcal{G}|} \sum_j^{|\mathcal{G}|} \Vert \text{diag}(\mathbf{S}_j) - \overline{\text{diag}(\mathbf{S}_j)} \cdot \mathbf{1}_{3\times1}\Vert_1
    \label{eq:mappingreg}
\end{equation}
%
%
The final mapping loss aggregates the aforementioned loss terms with corresponding weights as follows:
\begin{align}
    L_\text{map} = &\lambda_\text{map}\big(\lambda_\text{visual}L_\text{rgb} + (1 - \lambda_\text{visual})L_\text{dssim}\big) \label{eq:mappingtotloss} \\
                 & + (1 - \lambda_\text{map})L_\text{depth} + \lambda_\text{normals} L_\text{normals} + \lambda_\text{iso} L_\text{iso} \nonumber 
\end{align}
\vspace{-0.4cm}
\begin{table*}[!t]
    \centering
    \footnotesize
    \renewcommand{\tabcolsep}{12.4pt}
    \begin{tabular}{lcccccccc}
    \toprule
    Method                                                        & Kitchen        & Sofa           & Office         & Reception      & Living Room    & Office2        & Sofa2         &  Avg. \\
    \midrule
    KinectFusion~\cite{izadi2011kinectfusion}                     & \FAIL{}        & 0.146          & 0.209          & 0.157          & \FAIL{}        & 0.321          & 0.125         &  0.192       \\
    iMAP~\cite{sucar2021imap}                                     & \FAIL{}        & 1.658          & 0.338          & 0.648          & 0.679          & 0.344          & 0.214         &  0.647       \\
    NICE-SLAM~\cite{zhuNICESLAMNeuralImplicita}                   & 0.745          & 0.144          & 0.155          & 0.251          & 0.289          & 0.228          & 0.421         &  0.319       \\
    BundleFusion~\cite{daiBundleFusionRealtimeGlobally2017}       & \trd 0.176     & 0.102     & 0.135          & \trd 0.101     & \FAIL{}        & 0.163          &  0.120    &  0.132       \\
    ElasticFusion~\cite{whelanElasticFusionRealtimeDense2016}     & 0.253          & 0.110          &  0.070     & 0.193          & 0.530          & 0.121          & 0.146         &  0.203       \\
    MonoGS~\cite{monogs}                                          & 0.231          & \snd 0.032     & \snd 0.041     & \snd 0.044     & \snd 0.153     & \snd 0.035     & \snd 0.073    &  \snd 0.087  \\
    ToF-SLAM~\cite{tof-slam}                                      & \snd 0.113     & \trd 0.081     & \trd 0.056     & 0.114          & \trd 0.200     & \trd 0.101     & \trd 0.085    &  \trd  0.107  \\
    \textbf{\method{}  (ours)}                                    & \fst 0.088     & \fst 0.024     & \fst 0.022     & \fst 0.041     & \fst 0.122     & \fst 0.029     & \fst 0.030    &  \fst 0.051  \\
    \bottomrule
    \end{tabular}\vspace{-0.2cm}
    \caption{\textbf{ZJUL5 Tracking Results.} We show here the tracking performance on the ZJUL5 dataset~\cite{tof-slam} dataset with the ATE RMSE [m] (\da) metric on the 8 sequences available where \FAIL indicates lost tracking. \method provides the better performance by a good margin on each sequence, demonstrating the optimal capability of our proposal to track the camera ego-motion accurately.}\vspace{-0.3cm}
    \label{table:zjul5-trajectory}
\end{table*}

\section{Experiments}
\label{sec:experiments}

This section assesses both the tracking and mapping performance of \method on the following datasets.

\boldparagraph{ZJUL5.} ZJUL5~\cite{tof-slam} is the only existing public-domain real dataset providing sparse depth data from a low-resolution, low-cost ToF sensor. A VL53L5CX ToF sensor is assembled on a calibrated rig with an Intel RealSense 435i to provide dense depth ground truth. This benchmark comprises 7 diverse indoor scene recordings. Ground truth 3D meshes and camera poses are obtained following the ScanNet protocol \cite{dai2017scannet}. This sensor 
provides $8 \times 8$ depth points 
by counting the number of photons returned in each discretized time range. Of the at most 64 depth points provided, many are extremely noisy and introduce large outliers. 
This challenging real-world dataset provides real measurement noise that is not accurately represented in mainstream datasets.

\boldparagraph{TUM RGB-D.} This is a real dataset \cite{sturm2012benchmark} providing indoor sequence captures and widely used as benchmark for RGB-D SLAM pipelines. The depth stream provides dense VGA depth maps from a high-resolution Kinect v1 depth sensor.

\boldparagraph{Replica.} Replica \cite{replica19arxiv} is a synthetic dataset providing extremely realistic indoor sequences. It provides dense depth maps, which we use at different densities to perform experiments that prove the robustness and generalization capabilities of \method in challenging scenarios.

\boldparagraph{Baselines.} We compare our method with both learning-based \cite{sucar2021imap, zhuNICESLAMNeuralImplicita, tof-slam} and traditional pipelines \cite{izadi2011kinectfusion, whelanElasticFusionRealtimeDense2016, daiBundleFusionRealtimeGlobally2017}. Following ToF-SLAM~\cite{tof-slam}, baselines requiring RGB-D frames use DELTAR~\cite{deltar} to densify the sparse ToF measurements. Moreover, we adapt MonoGS~\cite{monogs} densifying depth with the state-of-the-art depth completion framework OGNI-DC~\cite{zuo2024ogni} for a Gaussian Splatting SLAM baseline.

\boldparagraph{Implementation Details.} In all our experiments, we use $N = 4$ source views $I_j$ (\emph{i.e.}, keyframe buffer elements) for the multi-view integration module and a baseline $b = 15\text{cm}$ for selecting keyframes. We use a quantile $q = 0.75$ to filter outliers that occur frequently in the sparse ToF depth of the ZJUL5 dataset \cite{tof-slam}. Concerning tracking, we use a novelty factor threshold of $\nu_{th} = 0.1,\ \sigma = 0.98$. The total tracking loss uses $\lambda_{\text{track}} = 0.9$, a total number of $\eta_\text{rgb}^T~=~30$ and $\eta_\text{rgbd}^T = 70$ steps are performed at each iteration. For mapping, we set the loss weights $\lambda_\text{map} = 60, \lambda_\text{visual} = 0.20, \lambda_\text{normals} = 0.01, \lambda_\text{iso} = 1.0$ and the number of steps per iteration $\eta^M_\text{rgbd} = 60$. We test on a single RTX3090.

\begin{figure}
    \centering
    \resizebox{\linewidth}{!}{
    \begin{tabular}{ccc}
    \raisebox{2.0\height}{\begin{turn}{90}\small\textbf Office\end{turn}} &
    \raisebox{0.0\height}{\includegraphics[trim=3.5cm 1cm 0 1cm, clip, width=0.8\linewidth]{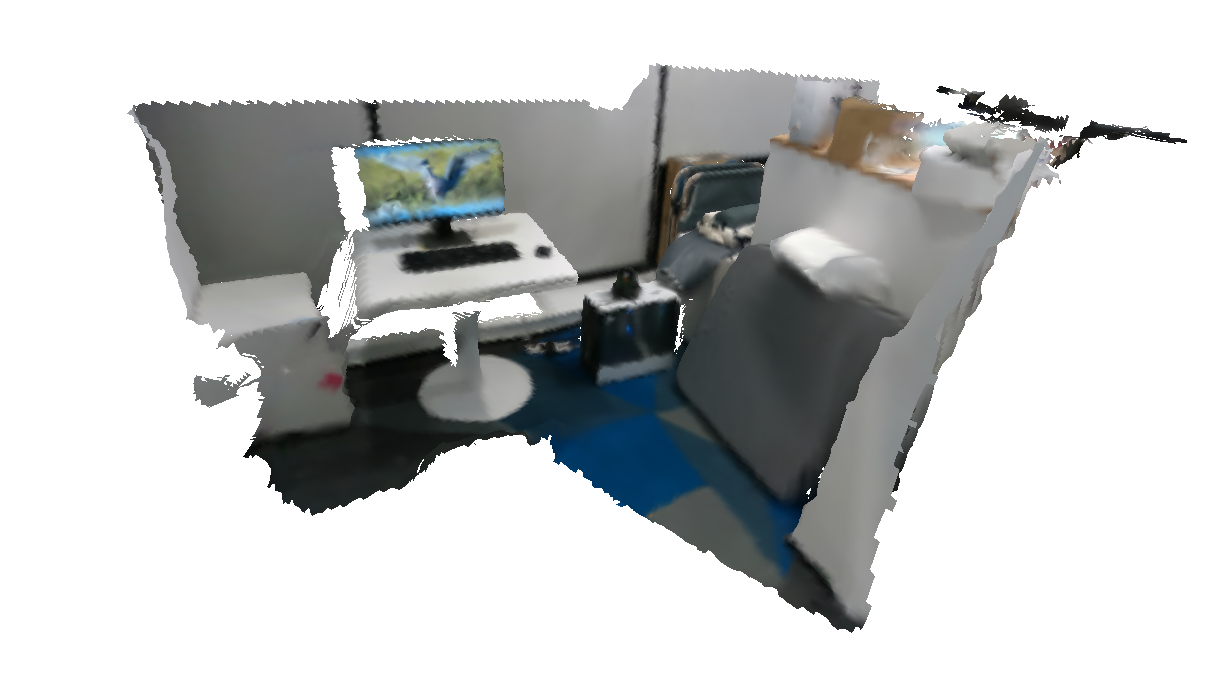}} &
    \includegraphics[trim=2cm 1cm 3cm 1cm, clip, width=0.55\linewidth]{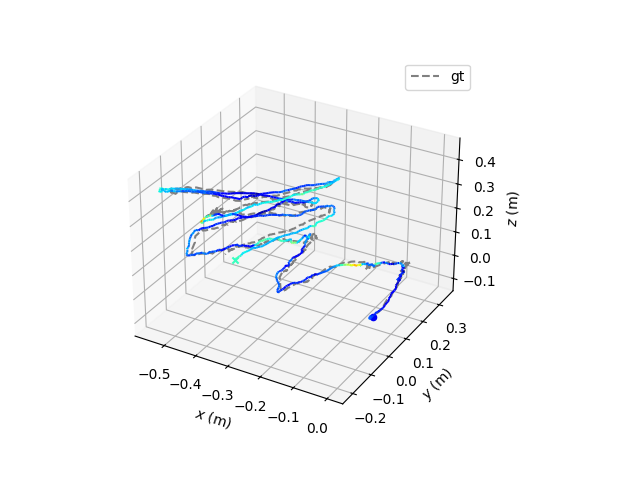} \\
    \raisebox{1.0\height}{\begin{turn}{90}\small\textbf Reception\end{turn}} &
    \raisebox{0.1\height}{\includegraphics[trim=3.5cm 1cm 0 1cm, clip, width=0.75\linewidth]{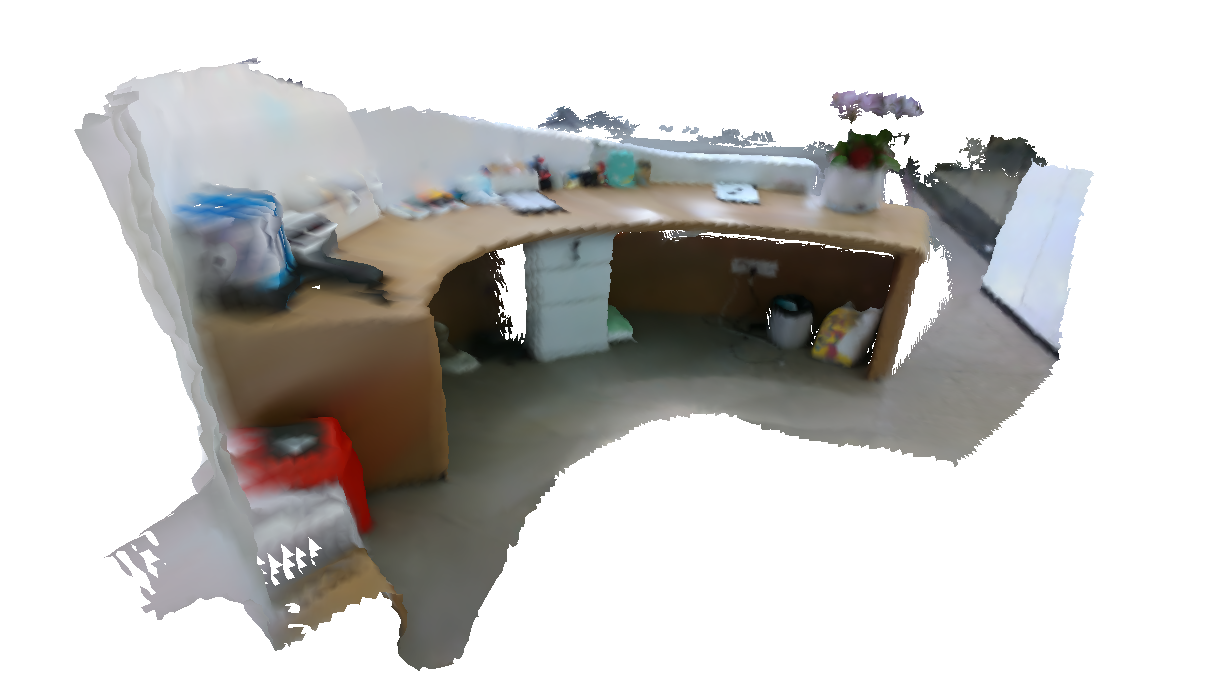}} &
    \includegraphics[trim=2cm 1cm 3cm 1cm, clip, width=0.55\linewidth]{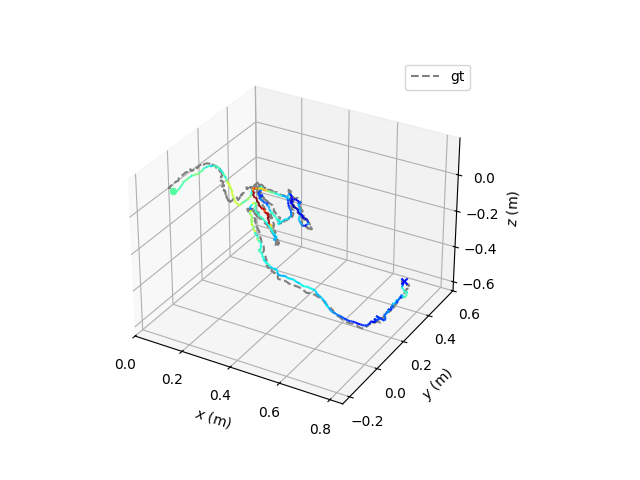} \\
    \raisebox{3.0\height}{\begin{turn}{90}\small\textbf Sofa\end{turn}} &
    \raisebox{0.05\height}{\includegraphics[trim=3.5cm 1cm 0 1cm, clip, width=0.75\linewidth]{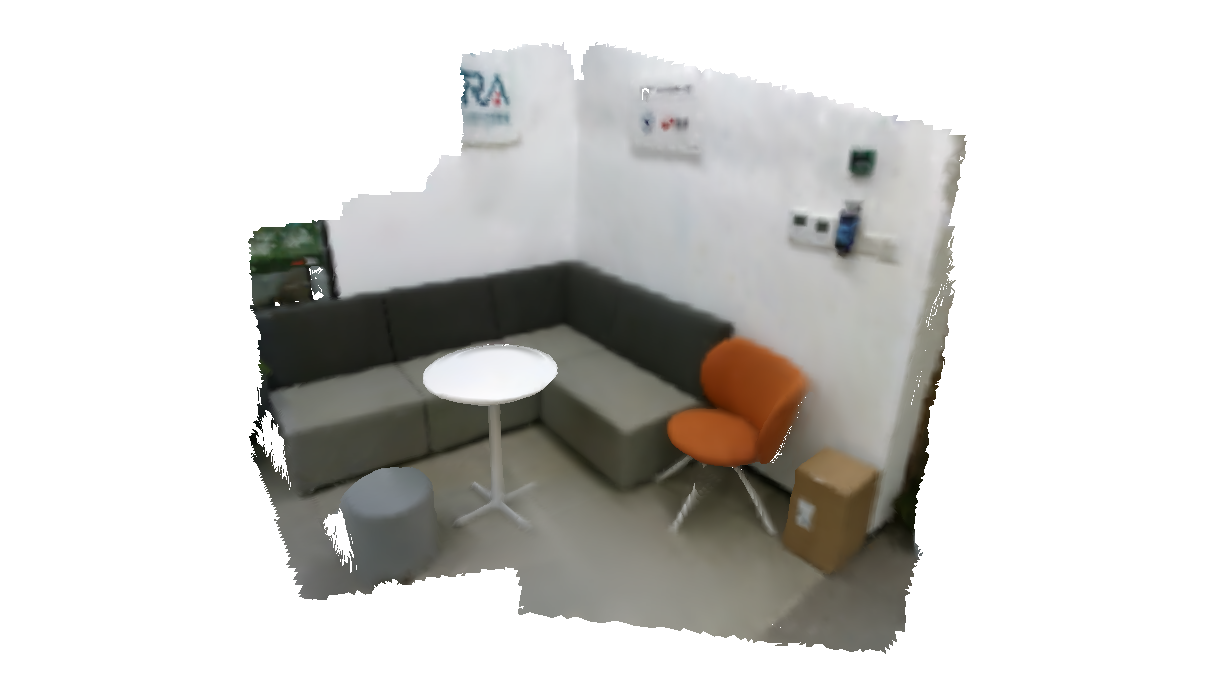}} &
    \includegraphics[trim=2cm 1cm 3cm 1cm, clip, width=0.55\linewidth]{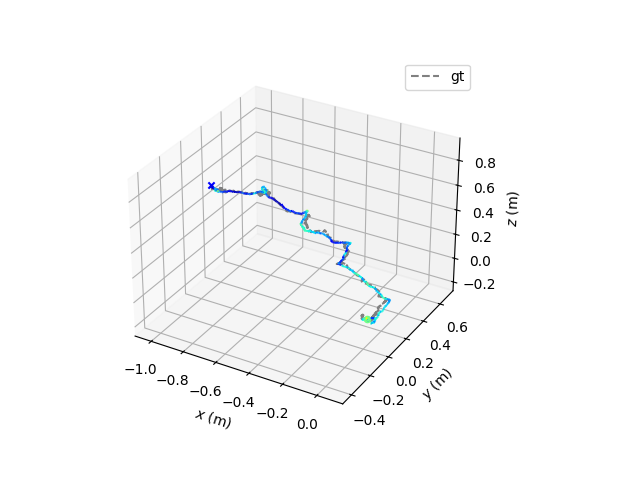} \\
    \multicolumn{3}{c}{
    \includegraphics[width=\linewidth]{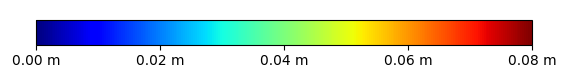}
    }
    \end{tabular}
    }\vspace{-0.2cm}
    \caption{\textbf{Qualitative results on the ZJUL5 dataset \cite{tof-slam}.} We show meshes obtained by fusing rendered depth maps with TSDF and marching cubes (left), and 3D trajectories (right) on 3 scenes selected from the ZJUL5 dataset \cite{tof-slam}.}\vspace{-0.3cm} 
    \label{fig:zjul5-qualitatives}
\end{figure}

\subsection{Comparison with the State of the Art}
In this section, we compare \method{} with existing methods in terms of tracking and mapping performances.


\boldparagraph{Tracking.} In Table~\ref{table:zjul5-trajectory}, we report the tracking performance by \method{} and various baselines on each sequence of the ZJUL5 dataset~\cite{tof-slam}. Following common practice  we use the Absolute Trajectory Error metric~\cite{sturm2012benchmark}, which directly compares the global consistency of a trajectory. While various frameworks~\cite{izadi2011kinectfusion, sucar2021imap, daiBundleFusionRealtimeGlobally2017} fail on the most challenging scenes, \method{} delivers accurate and consistent tracking. This is evident from the good margin it gains against the baselines. In Figure~\ref{fig:zjul5-qualitatives}, we show the 3D trajectories on a subset of the scenes, with error below $3$ cm. 

\begin{table*}[!t]
  \centering
  \footnotesize
  \setlength{\tabcolsep}{9pt}
  \begin{tabular}{llcccccccc}
  
    \toprule
    Method                         &  Metric         & Kitchen        & Sofa           & Office       & Reception      & Living Room     & Office2        & Sofa2          & Avg.           \\
    \hline
  
    \multirow{3}{*}{KinectFusion~\cite{izadi2011kinectfusion}}
                                   & Accuracy \da    & \FAIL         & 0.190          & 0.211        & 0.261          & \FAIL            & 0.267          & 0.135          & - \\ 
                                   & Completion \da  & \FAIL         & 0.048          & 0.046        & 0.064          & \FAIL            & \snd 0.078     & 0.064          & - \\ 
                                   & F-score \ua     & \FAIL         & 0.278          & 0.288        & 0.285          & \FAIL            & 0.274          & 0.381          & - \\ 
    \hline
    \multirow{3}{*}{ElasticFusion~\cite{whelanElasticFusionRealtimeDense2016}}
                                   & Accuracy \da    &      0.092    & 0.135          &      0.084   & 0.297          & 0.151            & \trd 0.096     & 0.122          &      0.140          \\
                                   & Completion \da  & \fst 0.065    & 0.048          & 0.082        & 0.305          & 0.216            & 0.147          & 0.047          & \trd 0.130          \\
                                   & F-score \ua     & \snd 0.553    & 0.420          & \trd 0.529   & 0.274          & \trd 0.382       & 0.416          & 0.481          & \trd 0.436          \\
    \hline
    \multirow{3}{*}{BundleFusion~\cite{daiBundleFusionRealtimeGlobally2017}}
                                   & Accuracy \da    & 0.170          &      0.100    & 0.103          &      0.122    & \FAIL           & 0.121          & 0.123          & - \\ 
                                   & Completion \da  & \trd 0.088     & \fst 0.030    & \snd 0.038     & \snd 0.057    & \FAIL           & 0.214          & \snd 0.034     & - \\ 
                                   & F-score \ua     & 0.373          & \trd 0.571    & 0.474          & \trd 0.470    & \FAIL           & \trd 0.442     & \trd 0.527     & - \\ 
    \hline
    \multirow{3}{*}{iMAP~\cite{sucar2021imap}}
                                   & Accuracy \da    & \FAIL          & 0.135          & 0.229         & 0.365          & 0.225          & 0.233          & 0.139          & - \\ 
                                   & Completion \da  & \FAIL          & 0.054          & 0.103         & 0.245          & 0.291          & 0.139          & 0.069          & - \\ 
                                   & F-score  \ua    & \FAIL          & 0.445          & 0.315         & 0.238          & 0.170          & 0.255          & 0.416          & - \\ 
    \hline
    \multirow{3}{*}{NICE-SLAM~\cite{zhuNICESLAMNeuralImplicita}}
                                   & Accuracy \da    & 0.303          & 0.119          & 0.116         & 0.216          &      0.103    & 0.156          & 0.464           & 0.211                 \\
                                   & Completion \da  & 0.456          & 0.042          & 0.070         & 0.199          & \fst 0.089    & 0.163          & 0.045           & 0.152                 \\
                                   & F-score \ua     & 0.221          & 0.554          & 0.411         & 0.402          & \snd 0.400    & 0.273          & 0.401           & 0.380                 \\
    \hline
    \multirow{3}{*}{MonoGS~\cite{monogs}}
                                   & Accuracy \da    &  \trd 0.090    &  \trd 0.086    &  \trd 0.083   & \snd 0.071     & \snd 0.069    & \snd 0.081     & \trd 0.089      & \snd 0.081            \\
                                   & Completion \da  &  0.246         &  0.122         &  0.104        & 0.238          & 0.193         & 0.157          & 0.170           & 0.175                 \\
                                   & F-score \ua     &  0.290         &  0.299         &  0.346        & 0.367          & 0.259         & 0.374          & 0.289           & 0.318                 \\
    \hline
    \multirow{3}{*}{ToF-SLAM~\cite{tof-slam}}
                                   & Accuracy \da    & \snd 0.081     & \snd 0.068     & \snd 0.067    & \trd 0.079     & \trd 0.078    &      0.113     & \snd 0.121      & \trd 0.087           \\
                                   & Completion \da  & \snd 0.071     & \trd 0.041     & \trd 0.045    & \trd 0.062     & \snd 0.122    & \trd 0.085     & \fst 0.033      & \snd 0.066           \\
                                   & F-score \ua     & \fst 0.559     & \snd 0.661     & \snd 0.646    & \snd 0.643     & \fst 0.496    & \snd 0.557     & \fst 0.656      & \snd 0.604           \\ 
    \hline
    \multirow{3}{*}{\textbf{\method{}  (ours)}}
                                   & Accuracy \da    & \fst 0.064     & \fst 0.031     & \fst 0.032    & \fst 0.041     & \fst 0.059    & \fst 0.034     & \fst 0.043      & \fst 0.043           \\
                                   & Completion \da  &  0.095         & \snd 0.038     & \fst 0.029    & \fst 0.056     & \trd 0.131    & \fst 0.054     & \trd 0.046      & \fst 0.064           \\
                                   & F-score \ua     &  \trd 0.527    & \fst 0.791     & \fst 0.840    & \fst 0.710     &      0.359    & \fst 0.779     & \snd 0.642      & \fst 0.664           \\
    \bottomrule
    \end{tabular}
    \vspace{-0.2cm}  
\caption{\textbf{ZJUL5 Mapping Results.} We perform the mapping evaluation on the ZJUL5 dataset (7 indoor sequences) and report results of three metrics including accuracy (Acc.), completion (Comp.), and F-score. The failure cases are marked as \FAIL.}\vspace{-0.20cm}
\label{table:zjul5-mapping}
\end{table*}
\newcommand{\twoc}[1]{\multicolumn{2}{c}{#1}}


\begin{table*}[!ht]
\centering
\footnotesize
\setlength{\tabcolsep}{7pt}
\begin{tabular}{cclcccccccc}
\toprule
Monocular Cues & Multi-View Cues & Metric & Kitchen    & Sofa & Office     & Reception  & Living Room & Office2    & Sofa2      & Avg.       \\ \hline

\multirow{2}{*}{\cmark} & \multirow{2}{*}{\xmark} & ATE $\downarrow$ & \snd 0.103 & \snd 0.054 & \trd 0.046 & \trd 0.050 & \snd 0.110 & \trd 0.041 & \trd 0.055 & \snd 0.066 \\
& & F-score $\uparrow$ & \snd 0.216 & \trd 0.675 & \trd 0.564 & \trd 0.645 & \snd \snd 0.241 & \trd 0.611 & \trd 0.557 & \trd 0.501 \\
\hline
\multirow{2}{*}{\xmark} & \multirow{2}{*}{\cmark} &  ATE $\downarrow$ & \trd 0.234 & \fst 0.024 & \snd 0.025 & \fst 0.037 & \fst 0.109 & \snd 0.034 & \fst 0.028 & \trd 0.072 \\
& & F-score $\uparrow$ & \trd 0.284 & \snd 0.781 & \snd 0.826 & \snd 0.703 & \trd 0.253 & \snd 0.766 & \fst 0.683 & \snd 0.614 \\
\hline
\multirow{2}{*}{\cmark} &  \multirow{2}{*}{\cmark} & ATE $\downarrow$ &  \fst 0.088 & \fst 0.024 & \fst 0.022 & \snd 0.041 & \snd 0.122 & \fst 0.029 & \snd 0.030 & \fst 0.051 \\
 &  & F-score $\uparrow$ & \fst 0.527 & \fst 0.791 & \fst 0.840 & \fst 0.710 & \fst 0.359 & \fst 0.779 & \snd 0.642 & \fst 0.664 \\
\bottomrule
\end{tabular}
\vspace{-0.2cm}
\caption{\textbf{Multi-Frame Integration Ablation Study.} Analysis of the impact of monocular and multi-view cues in the multi-frame integration step.
Monocular cues are less impactful than multi-view ones, boosting performance in the challenging scenes where multi-view cues struggle most. Nonetheless, their combination yields the best results.
}\vspace{-0.2cm}
\label{tab:components-ablation}
\end{table*}

\begin{table}[t]
    \centering
    \footnotesize
    \setlength{\tabcolsep}{2pt}
    \begin{tabular}{llll|cccc}
    \toprule
    Method    & LC & Input & Density & fr1/desk & fr2/xyz & fr3/office  & Avg. \\
    \hline
    DROID-VO & \xmark & RGB    & 0.00\% & 0.052 & 0.107 & 0.073 & 0.077 \\
    MonoGS   & \xmark & RGB    & 0.00\% & \trd 0.038 & \trd 0.046 & \trd 0.035 & \trd 0.040 \\
    \hline
    \multirow{2}{*}{\textbf{ToF-Splatting}} & \multirow{2}{*}{\xmark} & \multirow{2}{*}{RGB-D} & 0.02\% & \snd 0.030 & \snd 0.022 & \snd 0.030 & \snd 0.027 \\
    & & & 0.04\% & \fst 0.027 & \fst 0.019 & \fst 0.021 & \fst 0.022 \\
    \hline \hline
    MonoGS & \xmark & RGB-D & 100\% & 0.015 & 0.014 & 0.015 & 0.015 \\
    \hline  \hline
    ORB-SLAM2 & \cmark & RGB & 0.00\% & 0.019 & 0.006 & 0.024 & 0.016 \\
    ORB-SLAM2 & \cmark & RGB-D & 100\% & 0.016 & 0.040 & 0.010 & 0.010 \\
    \bottomrule
    \end{tabular}\vspace{-0.2cm}
    \caption{\textbf{Results on TUM RGB-D dataset.} Top: competitors w/o loop-closure (LC); middle: most representative competitor with $100\%$ density; bottom: competitors w/ loop-closure.}\vspace{-0.6cm}
    \label{table:tum}
\end{table}

\boldparagraph{Mapping.} In Table~\ref{table:zjul5-mapping}, we report the mapping performance by \method{} compared with the other baselines. For each scene, we collect the predicted pose of each frame and render depth and color from 3DGS. Then, we perform TSDF integration and extract the final mesh through marching cubes. Following \cite{tof-slam}, the meshes are then evaluated by computing Accuracy, Completion, and F-score versus the ground truth meshes. Accuracy and Completion respectively evaluate the mean distance between each predicted vertex and the nearest ground truth one and vice versa. The F-score takes into account both Accuracy and Completion with an aggregate metric. 
\method{} delivers consistent results for mapping, always achieving the highest accuracy and significantly exceeding the baselines on average and specifically on the Sofa, Office, and Reception scenes which account for five over seven sequences. \cite{tof-slam} is the second-best method and the main competitor in mapping. Notably, the other methods -- like \cite{whelanElasticFusionRealtimeDense2016, daiBundleFusionRealtimeGlobally2017} -- may provide good performance on specific scenes but demonstrate unreliability on average usually failing in the particularly challenging Kitchen and Living Room scenes. 
In Figure~\ref{fig:zjul5-qualitatives}, we show Office, Reception, and Sofa meshes on the left.

\begin{figure*}
    \centering
    \includegraphics[width=0.90\linewidth]{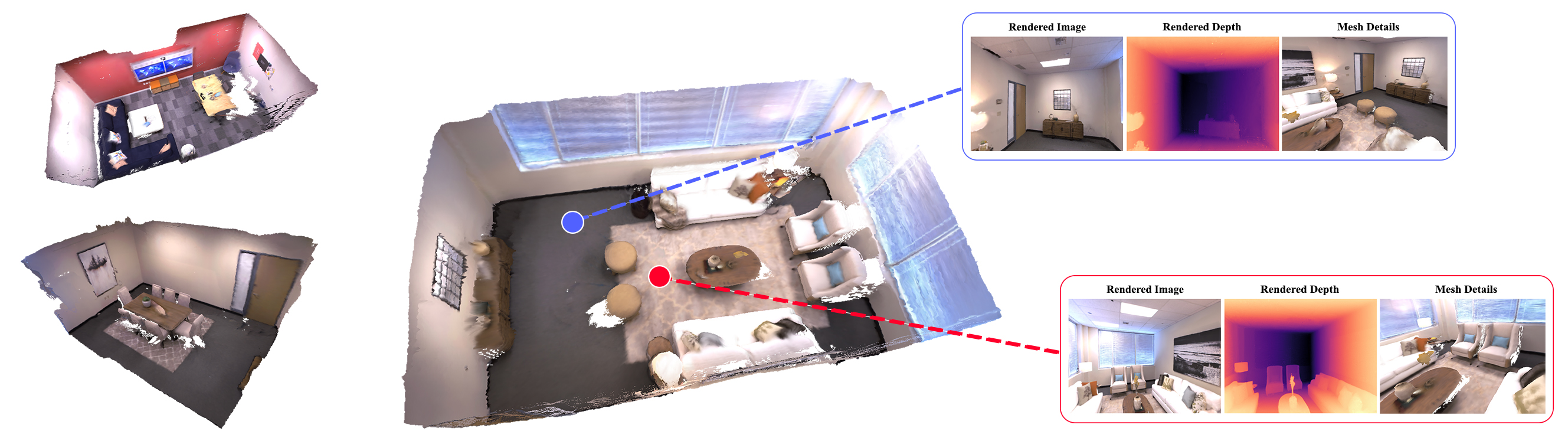}\vspace{-0.3cm}
    \caption{\textbf{Replica Qualitatives.} We provide qualitative results on Replica \cite{replica19arxiv} to demonstrate the generalization capabilities of our method. On the left from top to bottom, meshes obtained respectively on scenes Office2 and Room2. On the right, details from the scene Room0. \method{} delivers accurate details and allows for nice photometric and depth rendering.}
    \label{fig:replica-qualitative}
    \vspace{-0.3cm}
\end{figure*}

\begin{figure}
\centering
\begin{tikzpicture}
    \pgfplotsset{every tick label/.append style={font=\tiny}}
    \begin{groupplot}[
        group style={
            group size = 3 by 1,
            horizontal sep = 1.0cm,
        },
        axis lines = left,
        ymajorgrids=true,
        grid style=dashed,
        scaled ticks = false,
        scale only axis = true,
        xtick=data,
        yticklabel style={/pgf/number format/.cd, fixed, fixed zerofill, precision=2},
        width=3.5cm,
        height=2.0cm,
        xticklabel style={/pgf/number format/.cd, fixed, fixed zerofill, precision=2},
    ]

    \nextgroupplot[
        axis lines = left,
        title={\tiny \textbf{(a) MAE [m]}},
        grid style=dashed,
        ymin=0.030, ymax=0.055,
        height=1.5cm,
        width=1.5cm,
        xtick=\empty,
        extra x ticks={0.02, 0.04, 0.08, 0.16},
        extra x tick labels = {0.02\%, 0.04\%, 0.08\%, 0.16\%},
        extra x tick style={
            grid=major,
            tick label style={rotate=-90}
        }
    ]
    \addplot+[mark=triangle*] table[x=density,y=mae]{figures/data/density_ablation_ours.txt};

    \nextgroupplot[
        axis lines = left,
        title={\tiny \textbf{(b) PSNR [dB]}},
        grid style=dashed,
        scaled ticks = false,
        minor x tick num=1,
        ymin=26.8, ymax=28.3,
        height=1.5cm,
        width=1.5cm,
        scale only axis = true,
        yticklabel style={/pgf/number format/.cd, fixed, fixed zerofill, precision=1},
        xtick=\empty,
        extra x ticks={0.02, 0.04, 0.08, 0.16},
        extra x tick labels = {0.02\%, 0.04\%, 0.08\%, 0.16\%},
        extra x tick style={
            grid=major,
            tick label style={rotate=-90}
        },
    ]
    \addplot+[mark=triangle*] table[x=density,y=psnr]{figures/data/density_ablation_ours.txt};
    
    \nextgroupplot[
        axis lines = left,
        title={\tiny \textbf{(c) ATE [m]}},
        grid style=dashed,
        ymin=0.015, ymax=0.043,
        scaled ticks = false,
        minor x tick num=1,
        height=1.5cm,
        width=1.5cm,
        scale only axis = true,
        yticklabel style={/pgf/number format/.cd, fixed, fixed zerofill, precision=2},
        xtick=\empty,
        extra x ticks={0.02, 0.04, 0.08, 0.16},
        extra x tick labels = {0.02\%, 0.04\%, 0.08\%, 0.16\%},
        extra x tick style={
            grid=major,
            tick label style={rotate=-90}
        },
    ]
    \addplot+[mark=triangle*] table[x=density,y=ape]{figures/data/density_ablation_ours.txt};

    \end{groupplot}
\end{tikzpicture}\vspace{-2ex}
\caption{\textbf{Impact of depth sparsity.} We test on Replica~\cite{replica19arxiv} with different simulated depth sparsity levels to assess the capability to exploit higher input densities. MAE and ATE smoothly decrease, whereas rendering metrics appear to be less affected.}\vspace{-2ex}
\label{fig:density-study}
\end{figure}
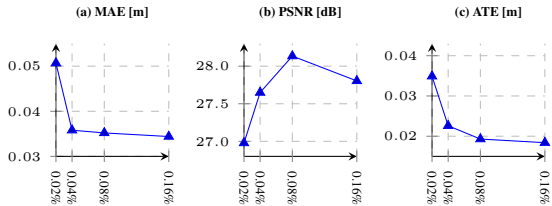
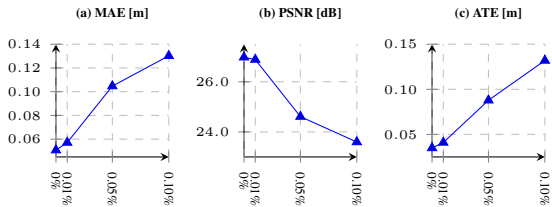
\begin{figure}
\centering
\begin{tikzpicture}
    \pgfplotsset{every tick label/.append style={font=\tiny}}
    \begin{groupplot}[
        group style={
            group size = 3 by 1,
            horizontal sep = 1.0cm,
        },
        axis lines = left,
        ymajorgrids=true,
        grid style=dashed,
        scaled ticks = false,
        scale only axis = true,
        xtick=data,
        yticklabel style={/pgf/number format/.cd, fixed, fixed zerofill, precision=2},
        width=3.5cm,
        height=2.0cm,
        xticklabel style={/pgf/number format/.cd, fixed, fixed zerofill, precision=2},
    ]

    \nextgroupplot[
        axis lines = left,
        ymin=0.045, ymax=0.140,
        title={\tiny \textbf{(a) MAE [m]}},
        grid style=dashed,
        height=1.5cm,
        width=1.5cm,
        xtick=\empty,
        extra x ticks={0.00, 0.01, 0.05, 0.10},
        extra x tick labels = {0\%, 0.01\%, 0.05\%, 0.10\%},
        extra x tick style={
            grid=major,
            tick label style={rotate=-90}
        },
    ]
    \addplot+[mark=triangle*] table[x=noise,y=mae]{figures/data/noise_ablation_ours.txt};

    \nextgroupplot[
        axis lines = left,
        title={\tiny \textbf{(b) PSNR [dB]}},
        grid style=dashed,
        ymin=23.0, ymax=27.5,
        scaled ticks = false,
        minor x tick num=1,
        height=1.5cm,
        width=1.5cm,
        scale only axis = true,
        yticklabel style={/pgf/number format/.cd, fixed, fixed zerofill, precision=1},
        xtick=\empty,
        extra x ticks={0.00, 0.01, 0.05, 0.10},
        extra x tick labels = {0\%, 0.01\%, 0.05\%, 0.10\%},
        extra x tick style={
            grid=major,
            tick label style={rotate=-90}
        },
    ]
    \addplot+[mark=triangle*] table[x=noise,y=psnr]{figures/data/noise_ablation_ours.txt};
    
    \nextgroupplot[
        axis lines = left,
        title={\tiny \textbf{(c) ATE [m]}},
        grid style=dashed,
        scaled ticks = false,
        ymin=0.025, ymax=0.150,
        minor x tick num=1,
        height=1.5cm,
        width=1.5cm,
        scale only axis = true,
        yticklabel style={/pgf/number format/.cd, fixed, fixed zerofill, precision=2},
        xtick=\empty,
        extra x ticks={0.00, 0.01, 0.05, 0.10},
        extra x tick labels = {0\%, 0.01\%, 0.05\%, 0.10\%},
        extra x tick style={
            grid=major,
            tick label style={rotate=-90}
        },
    ]
    \addplot+[mark=triangle*] table[x=noise,y=ape]{figures/data/noise_ablation_ours.txt};

    \end{groupplot}
\end{tikzpicture}\vspace{-2ex}
\caption{\textbf{Impact of depth noise.} We test on Replica \cite{replica19arxiv} injecting different amounts of noise $\xi \ \in \ [ 0.00, 0.01, 0.05, 0.10]$. \method{} demonstrates to be effective at dealing with noise, with error increasing almost linearly with the injected amount of noise.} 
\label{fig:noise-study}
\vspace{-0.5cm}
\end{figure}

\subsection{Ablation Studies}

In this section, we perform 
additional experiments aimed at studying the impact of different factors on \method.

\boldparagraph{Multi-Frame Integration.} In Table~\ref{tab:components-ablation}, we ablate the main components of the multi-frame integration and assess their contribution to the whole SLAM framework. By retaining monocular cues alone at the expense of multi-view geometry, we observe severe drops in both tracking and mapping accuracy, confirming the paramount importance of the latter; monocular cues alone are effective only where multi-view cues are ineffective -- \eg, in the absence of texture, as occurs mostly in the Kitchen scene.

\boldparagraph{TUM RGB-D.} In Table \ref{table:tum}, we simulate sparse ToF data by sampling 0.02\% and 0.04\% depth points, and compare \method{} with existing RGB and RGB-D methods. With as few as 0.02\% of points, \method{} largely improves over RGB methods \cite{teed2021droid,monogs}, approaching the 100\% density RGB-D approach of the most representative competitor \cite{monogs} at only 0.04\% density.

\boldparagraph{Replica: Input Depth Sparsity.} 
We conduct additional studies on the Replica dataset, for which Figure~\ref{fig:replica-qualitative} provides a qualitative overview of our 3D reconstructions. 
\method{} enables performing SLAM with very few sparse points -- the 64 sparse points provided by a $8 \times 8$ sparse account only for 0.02\% of a $640 \times 480$ image. We here explore the impact of higher sparse points density on its overall performance. We simulate on Replica densities of $\{0.02\%,\, 0.04\%, \, 0.08\%, \, 0.16\%\}$ corresponding to about $\{64, 128, 256, 512\}$ points. 
Figure~\ref{fig:density-study} shows the performance variation on both tracking and mapping metrics using the aforementioned sparsities.

\boldparagraph{Replica: Input Depth Noise.} We study various noise levels impact in the input depth to assess our pipeline robustness. We model noise as additive but dependent on depth, injecting heteroscedastic Gaussian noise $\mathcal{N}(d, \epsilon d)$ where $\epsilon$ modulates the variance of the noise source on Replica. Figure~\ref{fig:noise-study} shows that MAE, PSNR, and ATE all expose linear degradation trends as $\epsilon$ increases.

\boldparagraph{Runtime.} In Table~\ref{table:runtime} we measure the runtime of \method{} in comparison with the main baselines. The tracking time of \method{} already includes depth estimates with DoD, with average runtime $\bar{t}_{\rm DoD} \approx 90 \, ms$ (comprising DepthAnything v2 for monocular cues inference~\cite{depth-anything-v2}). To date, 3DGS-based mapping methods are not capable of real-time performances. Nevertheless, exploiting the fact that the mapping step accounts only for just a few frames (typically 4\% of the total in sequences of the ZJUL5 dataset~\cite{tof-slam}) and can be parallelized, the pipeline achieves $1.5$ fps without further optimization.

\begin{table}[!t]
\centering
\footnotesize
\setlength{\tabcolsep}{12pt}
\begin{tabular}{lrr}
\toprule
Method Name    & Tracking Time & Mapping Time \\ \hline
iMAP~\cite{sucar2021imap}          & 101 ms   & 448 ms  \\
NICE-SLAM~\cite{zhuNICESLAMNeuralImplicita}     & 470 ms   & 1300 ms \\
MonoGS~\cite{monogs}  &  1169 ms                 &  650 ms                    \\  
ToF-SLAM~\cite{tof-slam}        & 116 ms   & 380 ms  \\
\textbf{\method{} (ours)}  & 614 ms   & 865 ms          \\
\bottomrule
\end{tabular}
\vspace{-0.2cm}
\caption{\textbf{Runtime Comparison.} We compare the runtime of \method{} and other learned baselines measuring their runtime for tracking and mapping. MonoGS \cite{monogs} tracking time contains also the depth completion inference time \cite{zuo2024ogni}.}
\label{table:runtime}
\vspace{-0.3cm}
\end{table}

\boldparagraph{Limitations.} Even with state-of-the-art quality, \method manifests some issues that will be addressed in future developments. The most concerning is the runtime, which is too slow for real-time applications and requires memory and compute-intensive backpropagation at deployment. Even though fast 3DGS frameworks exist, such methodologies have not been transferred yet into the SLAM community, that being outside the scope of this work.
\section{Conclusion}
We presented the first 3DGS-based SLAM pipeline relying on sparse ToF depth sensing, as it provides accurate tracking and mapping from cheap and low-power cameras. Moreover, we showed how this result can be achieved through the integration of multi-view geometry, sparse depth data, and monocular cues, yielding an end-to-end 3DGS-based SLAM system. Finally, we assessed the effectiveness of our approach on the ZJUL5 and Replica datasets.

{
    \small
    \bibliographystyle{ieeenat_fullname}
    \bibliography{main}
}

\clearpage
\setcounter{page}{1}
\maketitlesupplementary


We provide this manuscript as a supplementary resource to the CVPR submission \#6364 titled ``ToF-Splatting: Dense SLAM using Sparse Time-of-Flight Depth and Multi-Frame Integration'' to provide a deeper understanding of the proposed framework through extended insights, detailed explanations, and additional qualitative results that complement the findings presented in the main paper. By including these extended materials, we hope to facilitate a more comprehensive appreciation of the contributions and practical relevance of the proposed approach.

\section{TUM RGB-D DoD Qualitative Results}

\begin{figure}[t]
    \newcommand{\sz}{1.95cm}
    \centering
    \null\hfill
    \includegraphics[height=\sz,trim={0.3cm 0.3cm 0.3cm 0.3cm},clip]{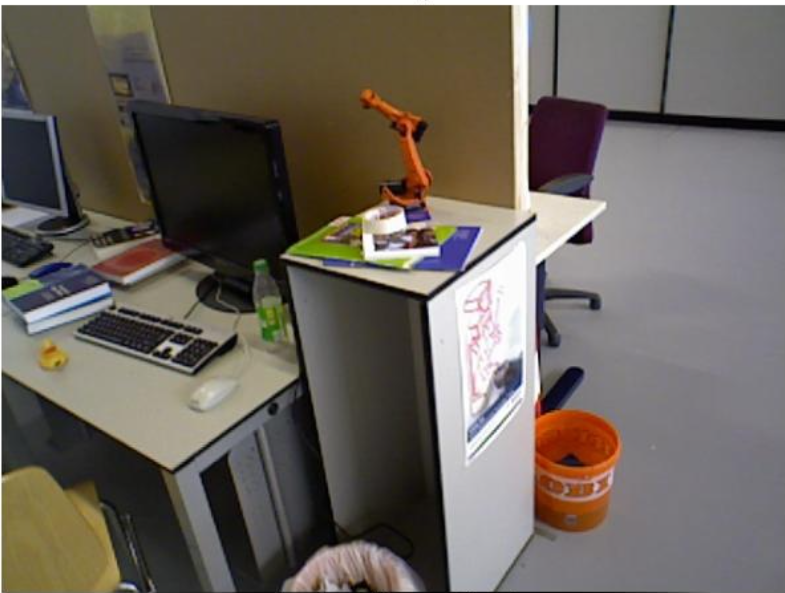}
    \hfill
    \includegraphics[height=\sz, trim={0.3cm 0.3cm 0.3cm 0.3cm},clip]{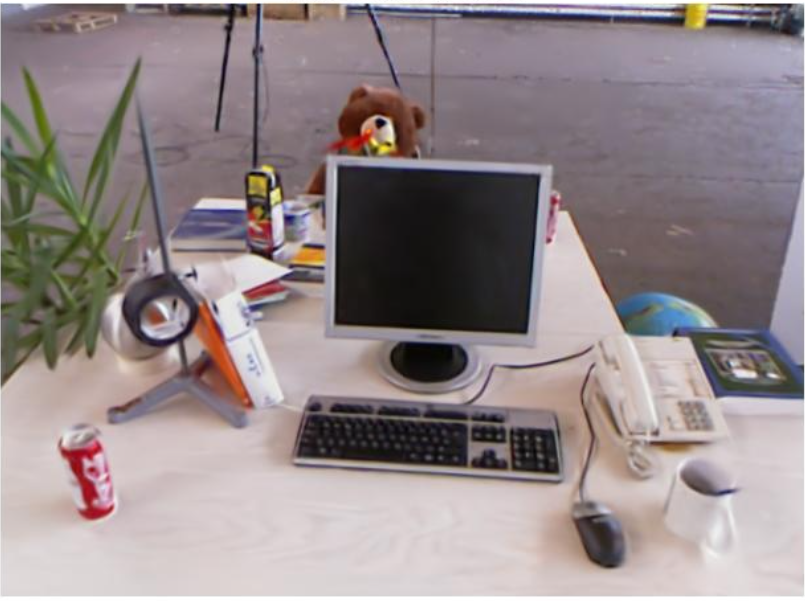}
    \hfill
    \includegraphics[height=\sz, trim={0.3cm 0.3cm 0.3cm 0.3cm},clip]{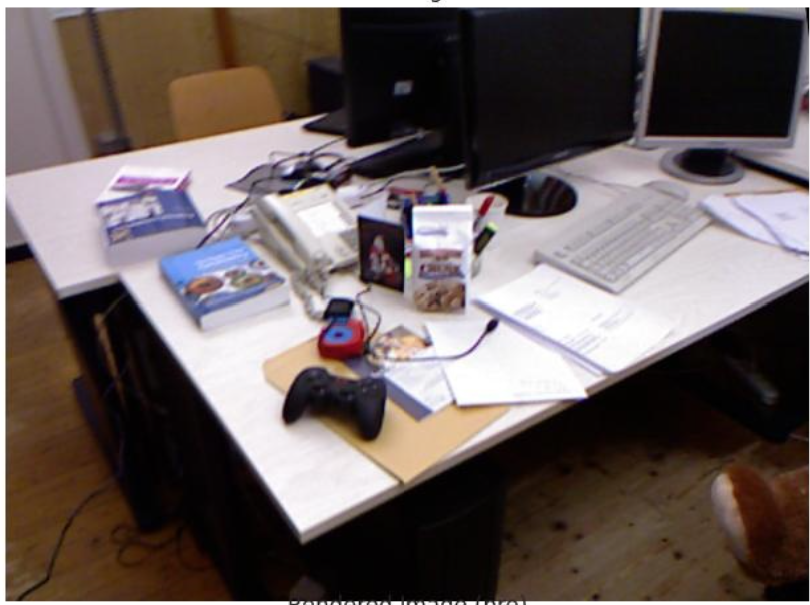} \hfill\null \\[-0.35cm]
    \null\hfill\includegraphics[height=\sz, trim={0.3cm 0.3cm 0.3cm 0.3cm},clip]{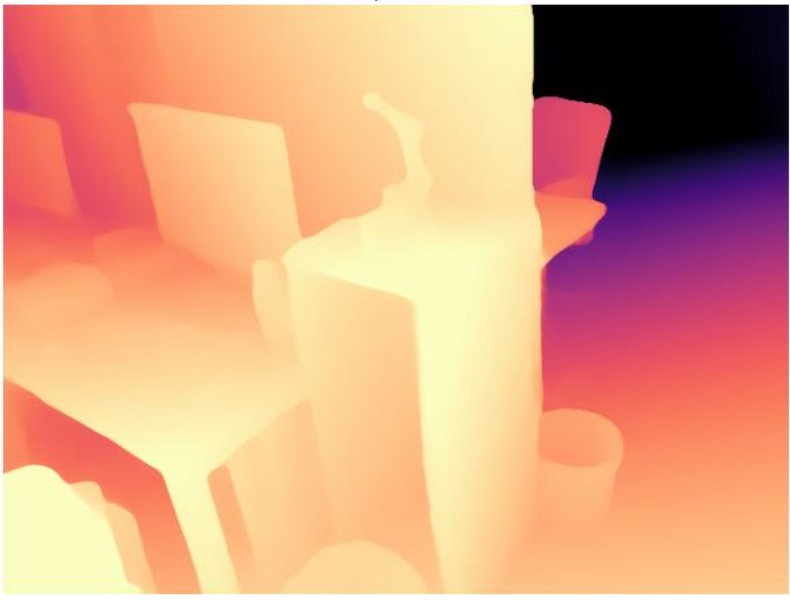}
    \hfill
    \includegraphics[height=\sz, trim={0.3cm 0.3cm 0.3cm 0.3cm},clip]{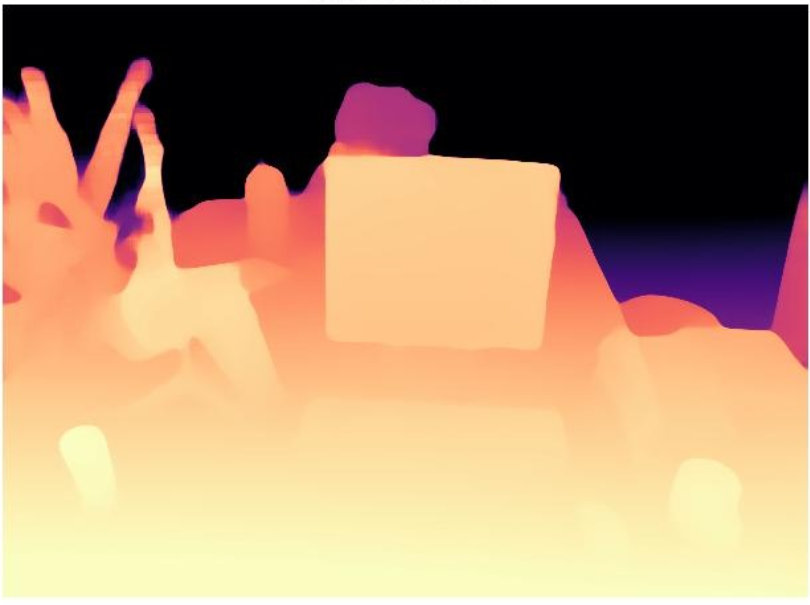}
    \hfill
    \includegraphics[height=\sz, trim={0.3cm 0.3cm 0.3cm 0.3cm},clip]{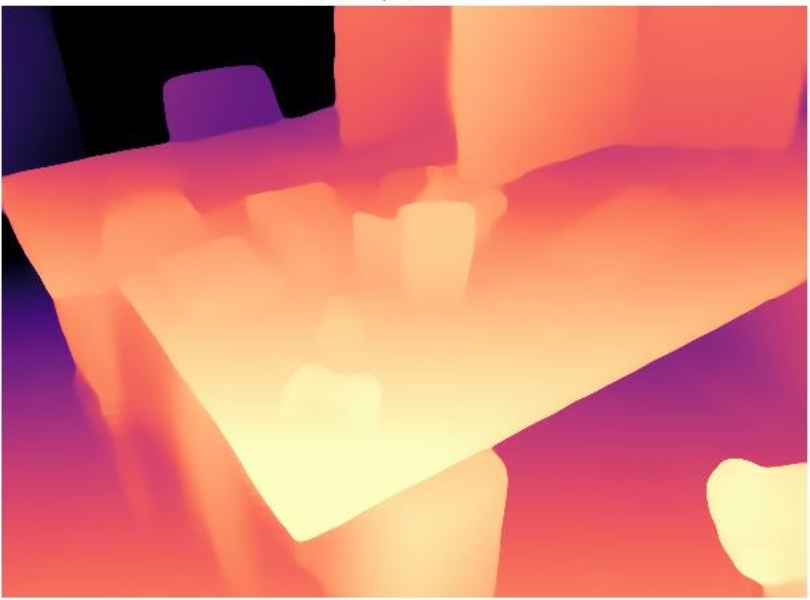}
    \hfill\null
    \caption{\textbf{Qualitative results on TUM RGB-D dataset.} Depth maps for fr1/office (left), fr2/xyz (center), fr3/desk (right) as estimated by DoD~\cite{depth-on-demand}.}
    \label{fig:tum-samples}
\end{figure}

In Figure~\ref{fig:tum-samples} we present three pairs of color views and reconstructed depth maps to assess that our method~\cite{depth-on-demand}, as integrated in the ToF-splatting pipeline, generalizes well to unseen datasets such as TUM RGB-D at test time. Indeed, we obtain high-quality depth maps from sparse inputs.

\section{Replica Qualitative Results}

\begin{figure*}
    \centering
    \begin{tabular}{@{}cccc@{}}
    \multicolumn{2}{c}{\scriptsize{Office 0}} & \multicolumn{2}{c}{\scriptsize{Office 1}} \\
    \includegraphics[width=0.22\linewidth]{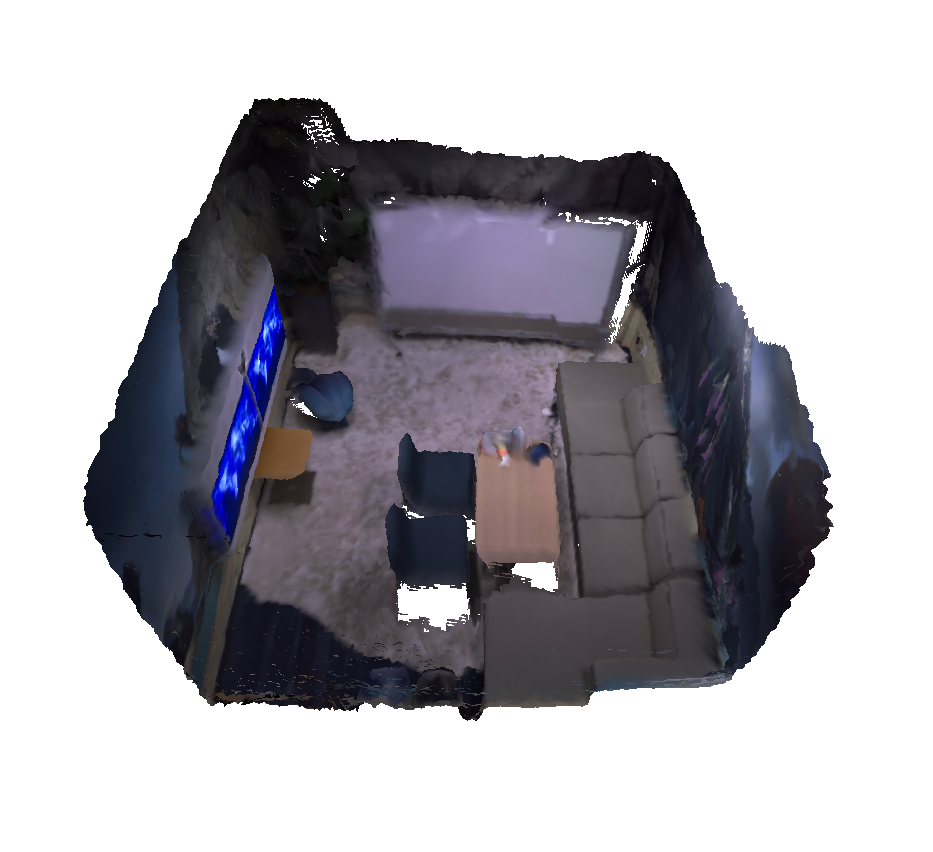} &
    \includegraphics[width=0.22\linewidth]{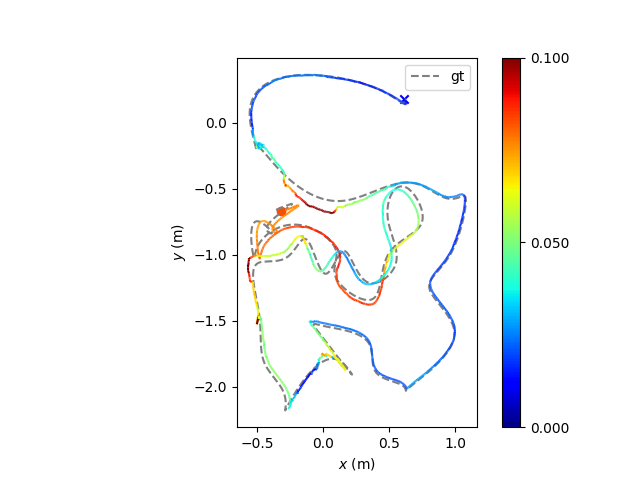} &
    \includegraphics[width=0.22\linewidth]{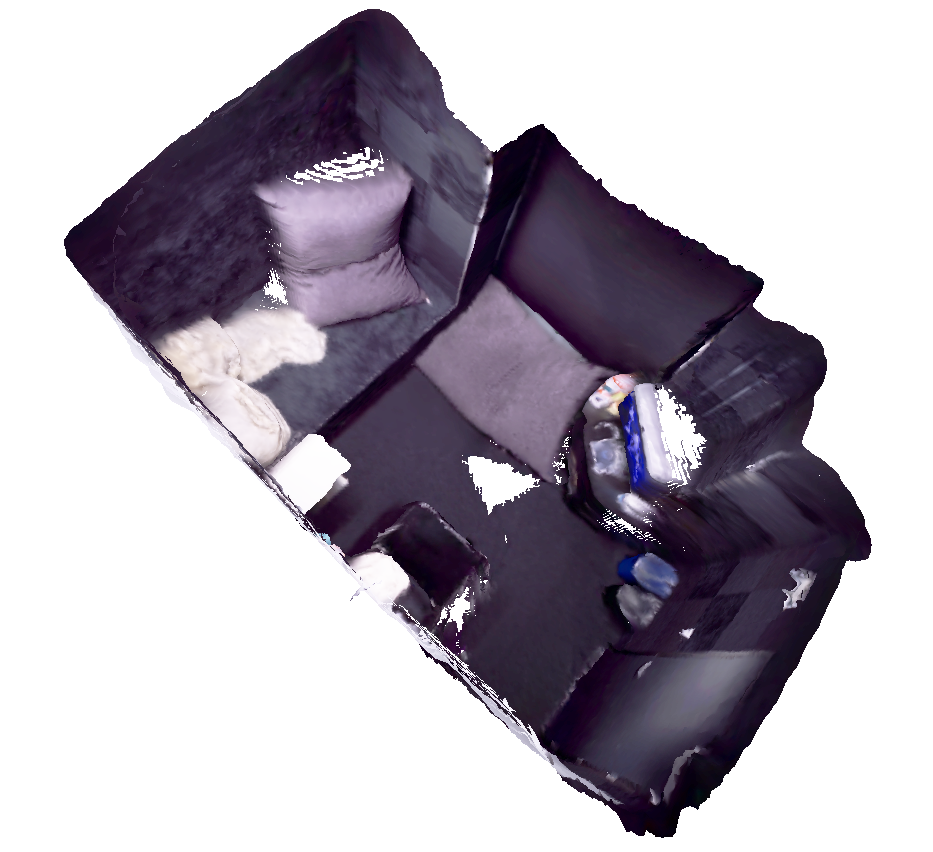} &
    \includegraphics[width=0.22\linewidth]{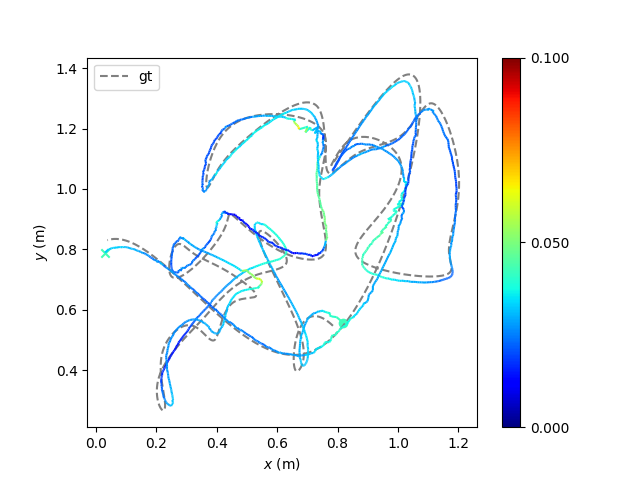} \\
    \multicolumn{2}{c}{\scriptsize{Office 2}} & \multicolumn{2}{c}{\scriptsize{Office 3}} \\
    \includegraphics[width=0.22\linewidth]{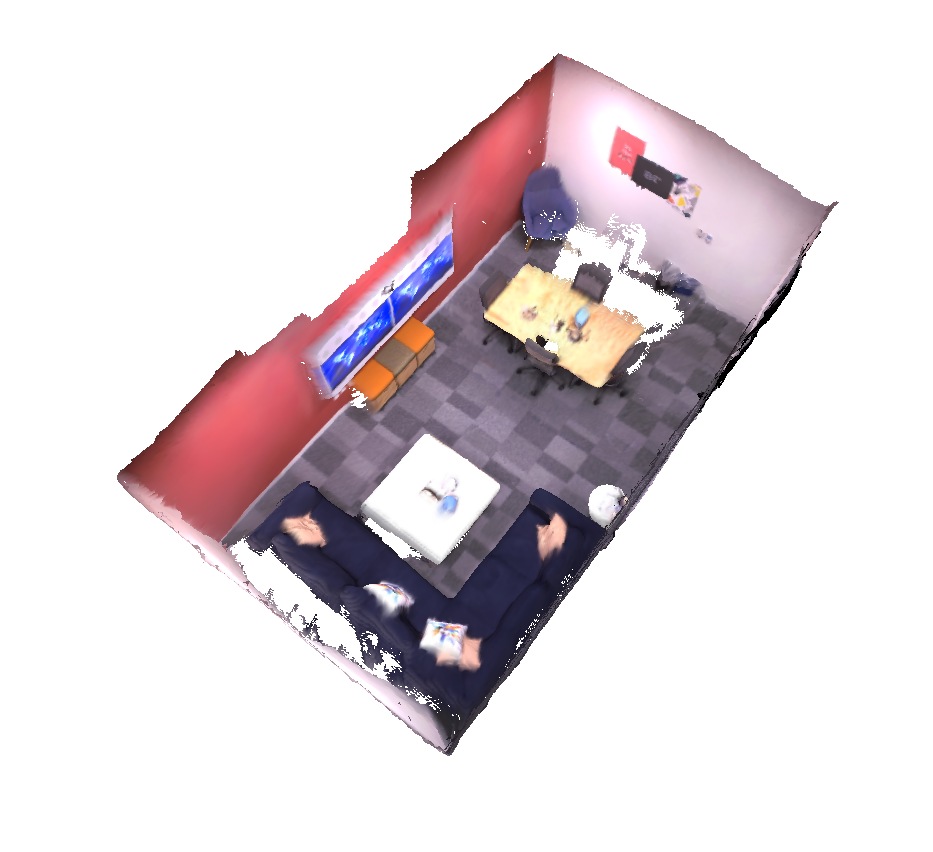} &
    \includegraphics[width=0.22\linewidth]{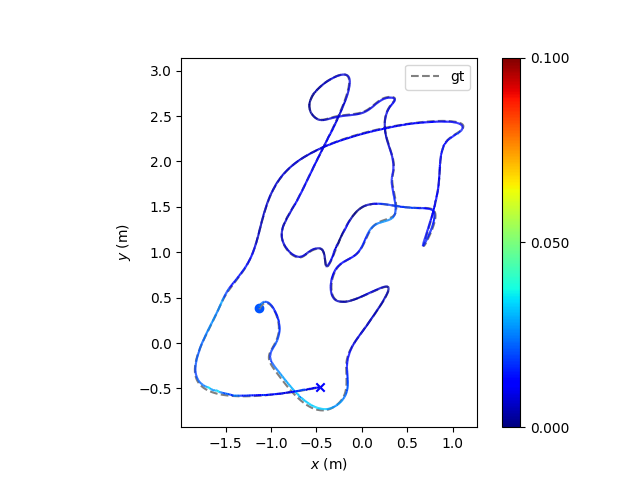} &
    \includegraphics[width=0.22\linewidth]{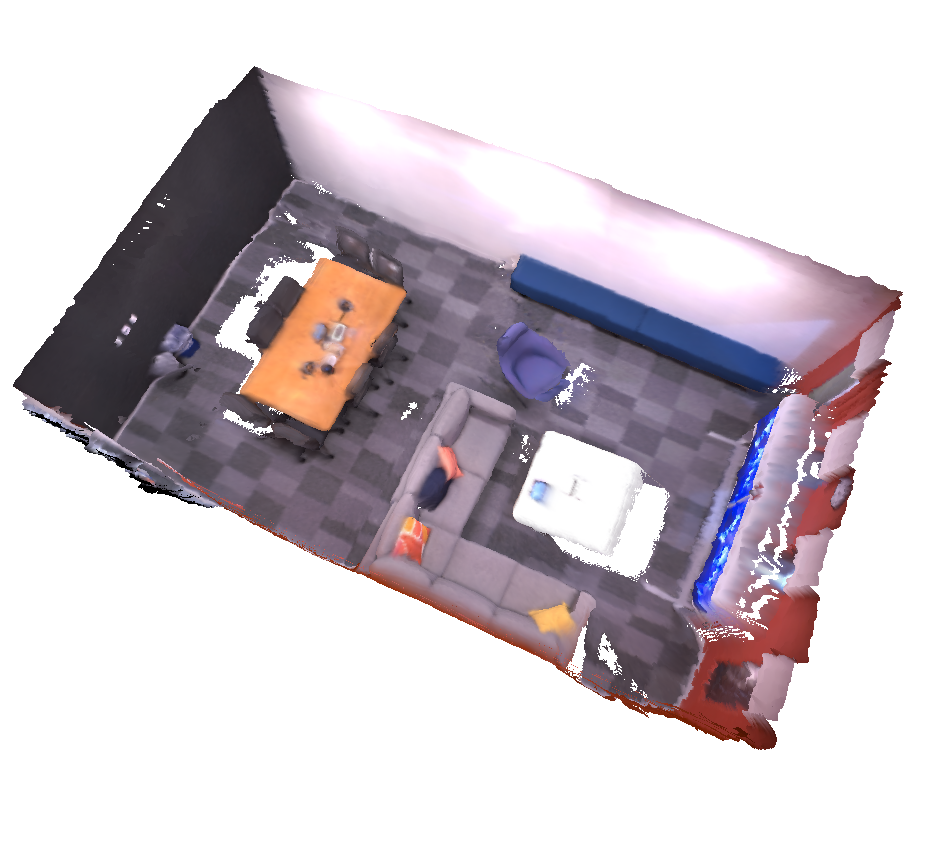} &
    \includegraphics[width=0.22\linewidth]{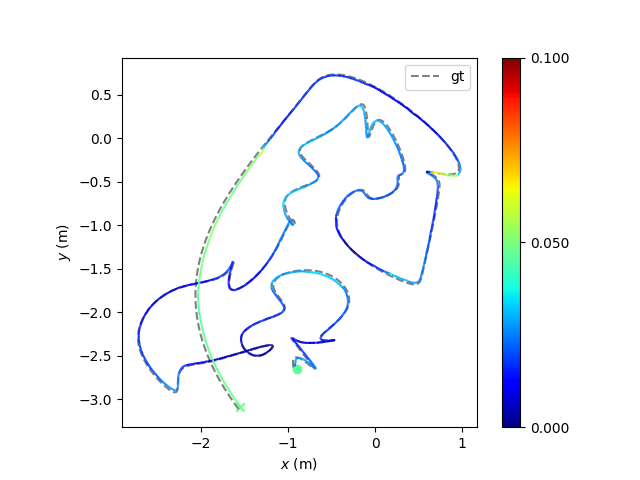} \\
    \multicolumn{2}{c}{\scriptsize{Office 4}} & \multicolumn{2}{c}{\scriptsize{Room 0}} \\
    \includegraphics[width=0.22\linewidth]{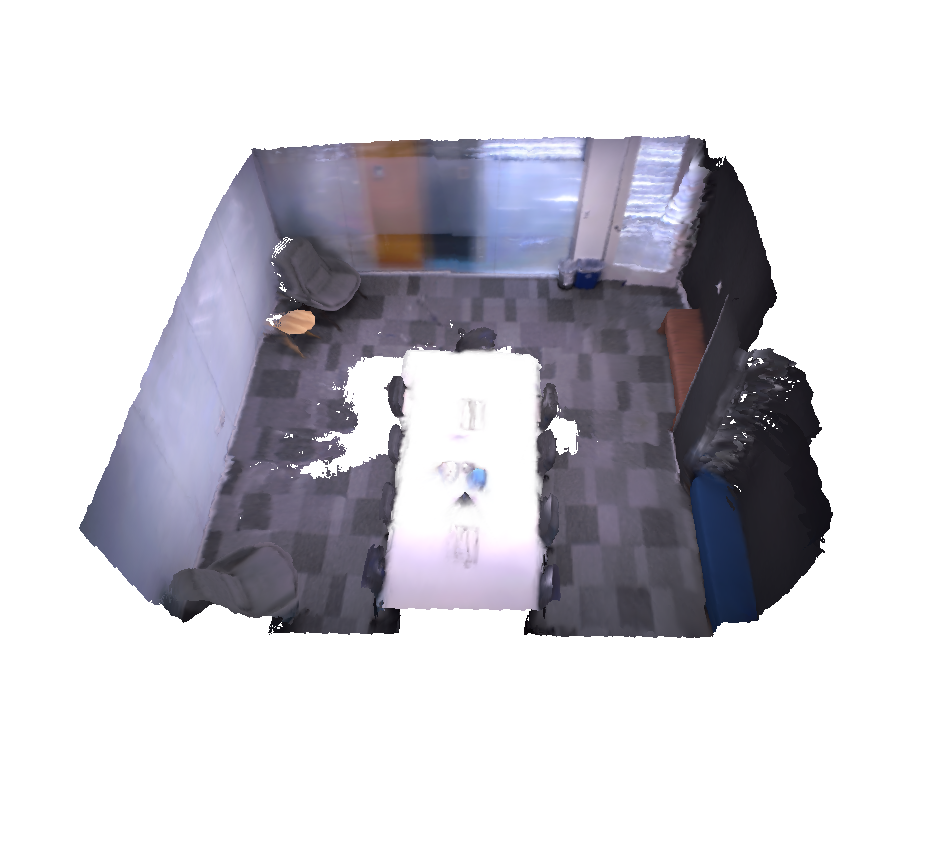} &
    \includegraphics[width=0.22\linewidth]{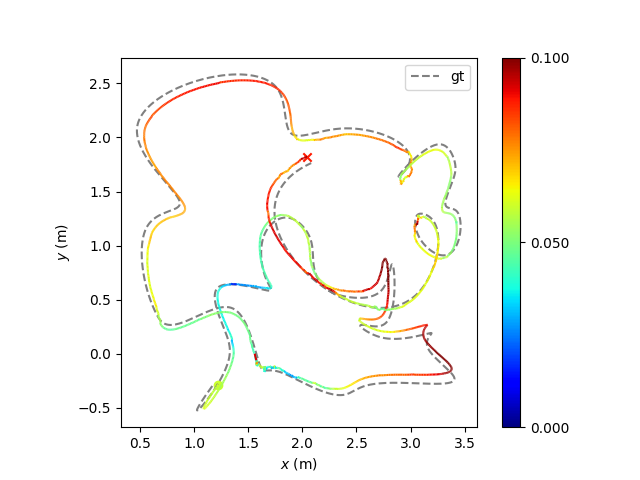} &
    \includegraphics[width=0.22\linewidth]{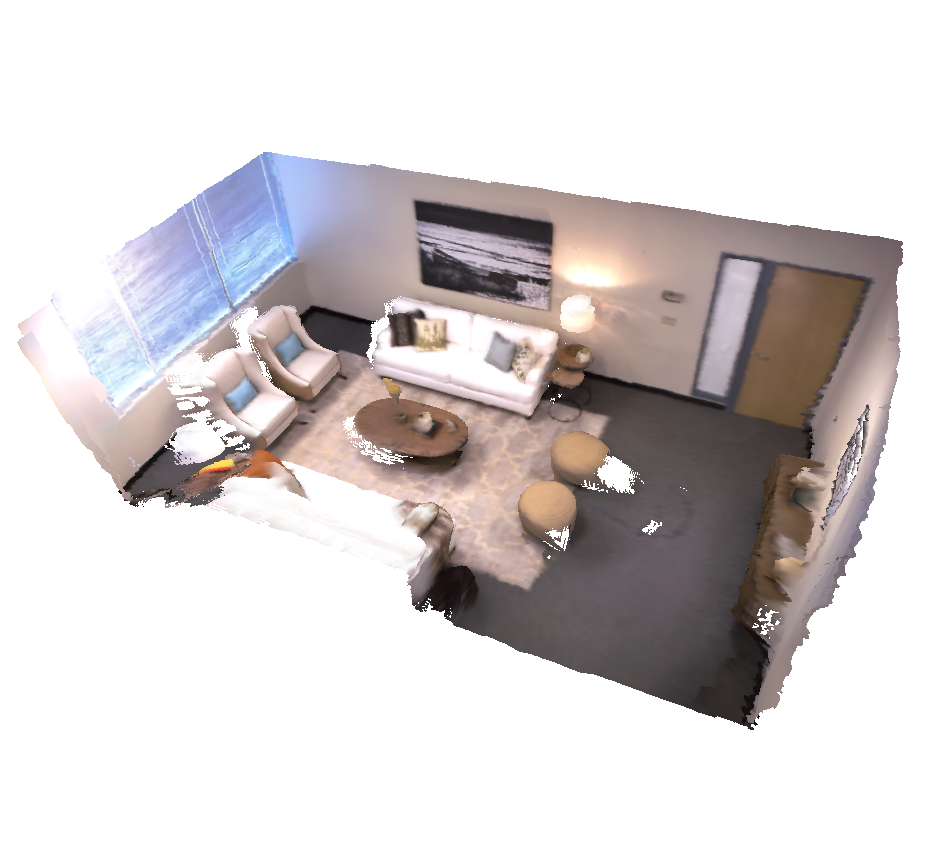} &
    \includegraphics[width=0.22\linewidth]{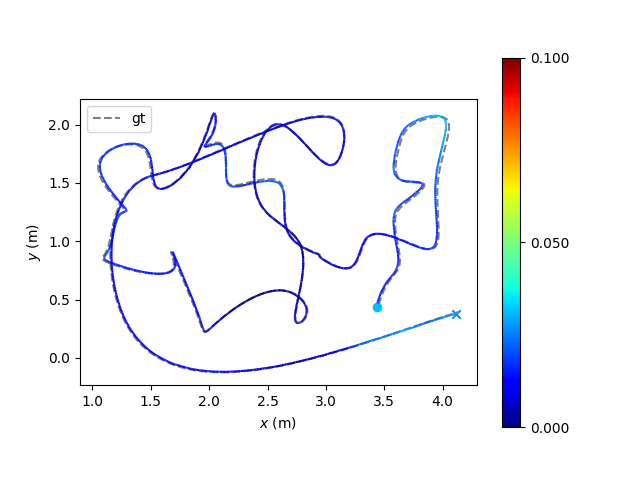} \\
    \multicolumn{2}{c}{\scriptsize{Room 1}} & \multicolumn{2}{c}{\scriptsize{Room 2}} \\
    \includegraphics[width=0.22\linewidth]{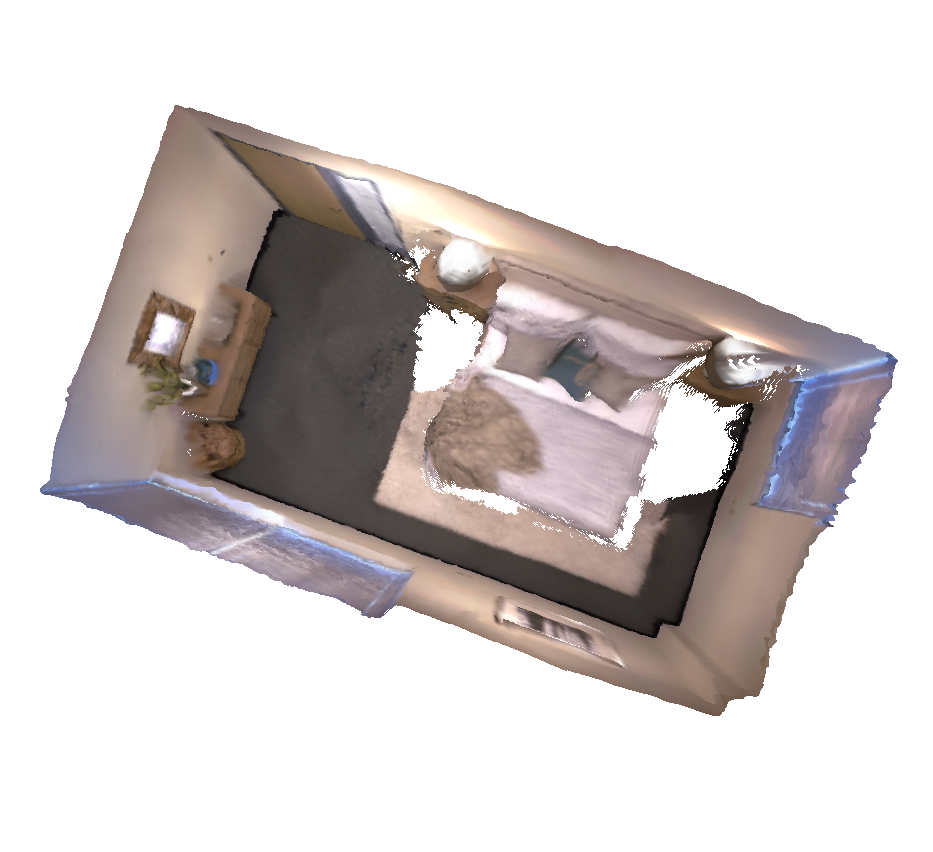} &
    \includegraphics[width=0.22\linewidth]{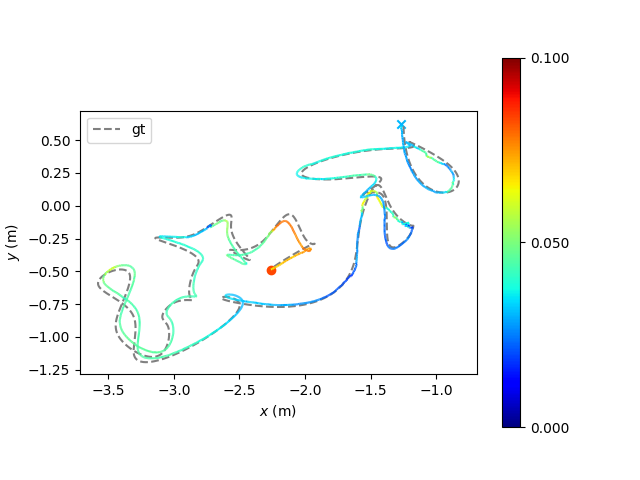} &
    \includegraphics[width=0.22\linewidth]{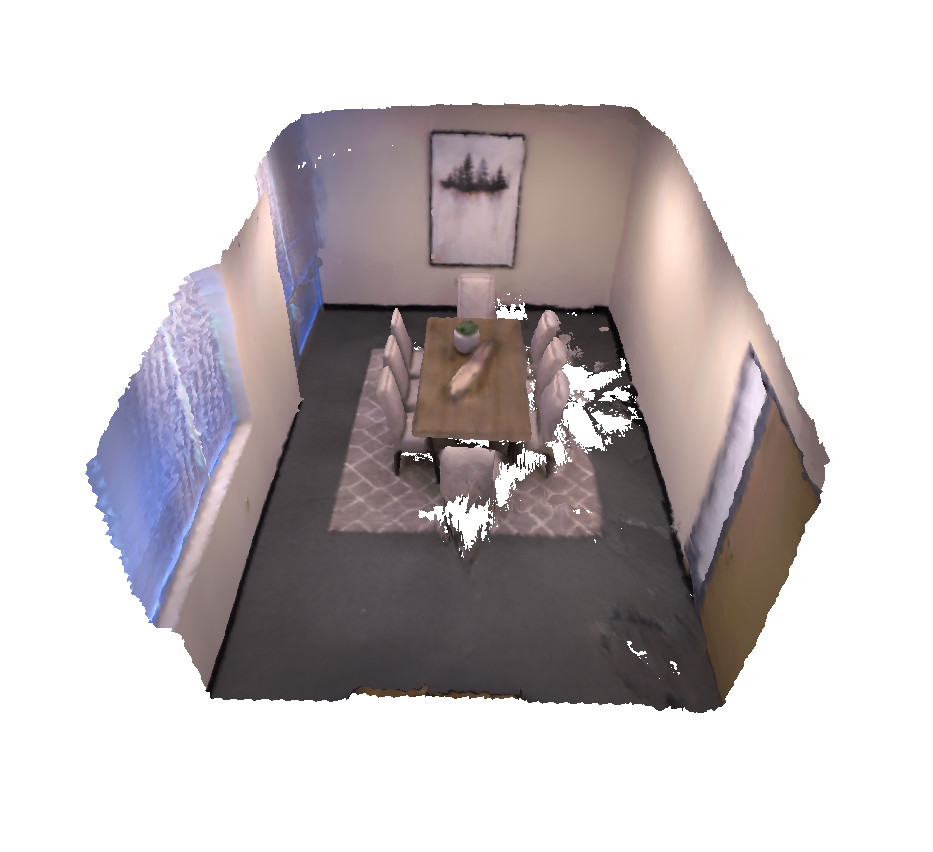} &
    \includegraphics[width=0.22\linewidth]{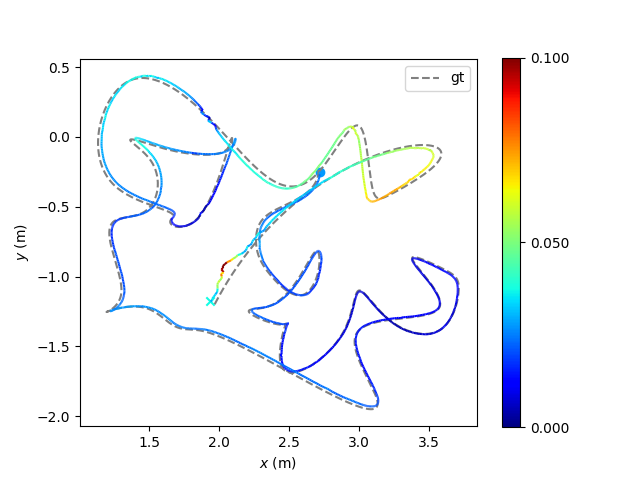} \\
    \end{tabular}
    \caption{\textbf{Replica Qualitatives.} We provide the trajectory and mesh reconstruction qualitative results on each scene provided by Replica. \method enables effective mesh reconstructions and accurate tracking.}
    \label{fig:replica-supplementary-qualitatives}
\end{figure*}

In Figure~\ref{fig:replica-supplementary-qualitatives}, we present the reconstructed mesh and the predicted trajectory for each scene in the Replica \cite{replica19arxiv} dataset. To achieve this, we first fit the entire scene and render depth and color images for each pose estimated by \method. These rendered outputs are then fused using Truncated Signed Distance Function (TSDF) integration. Once the integration is complete, we extract the 3D mesh using the marching cubes algorithm as implemented in Open3D. The reconstruction process employs a voxel size of 0.02 meters, with depth values truncated at a maximum distance of 4 meters to ensure robustness. On the right side of the figure, we visualize the $xy$ plane projections of the predicted and ground truth trajectories, where the ground truth is represented by a dashed gray line. Additionally, a color bar indicates the positional error between the corresponding frame poses along the trajectories. The results demonstrate that \method effectively achieves high-quality reconstructions and accurate tracking, highlighting its robustness and precision in this context.

\section{ZJUL5 Qualitative Results}

\begin{figure*}
    \centering
    \begin{tabular}{@{}cccc@{}}
    \multicolumn{2}{c}{\scriptsize{Reception}} & \multicolumn{2}{c}{\scriptsize{Kitchen}} \\
    \includegraphics[width=0.22\linewidth]{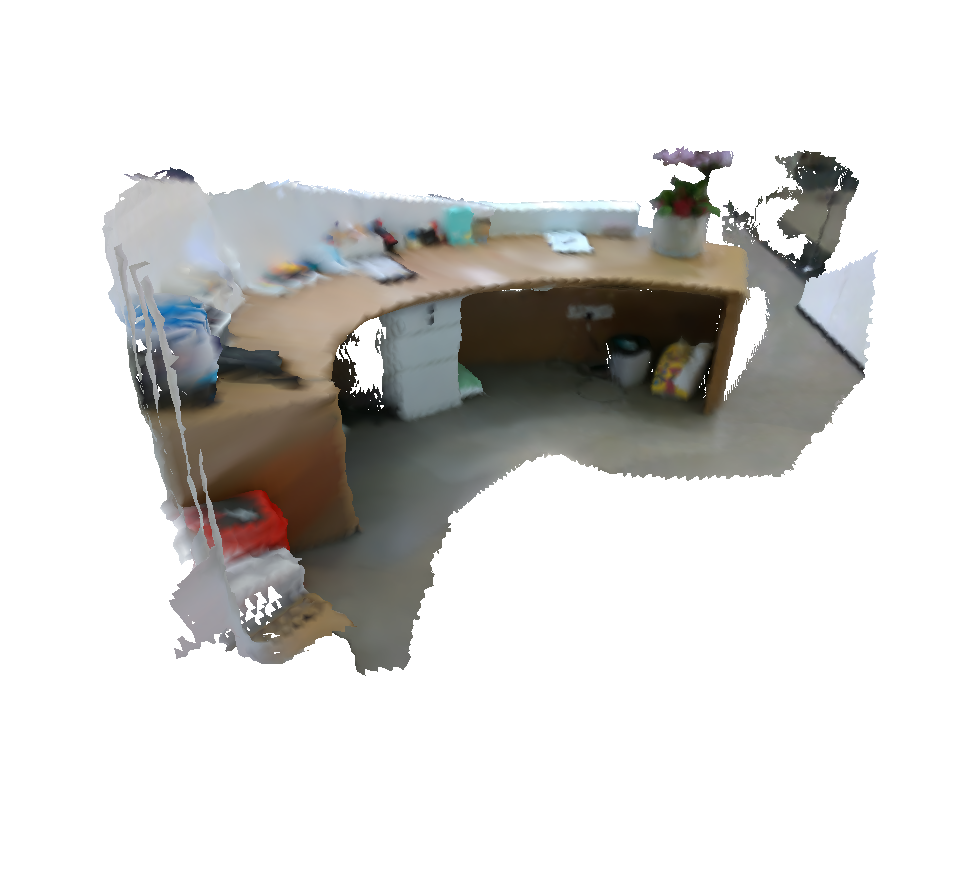} &
    \includegraphics[width=0.22\linewidth]{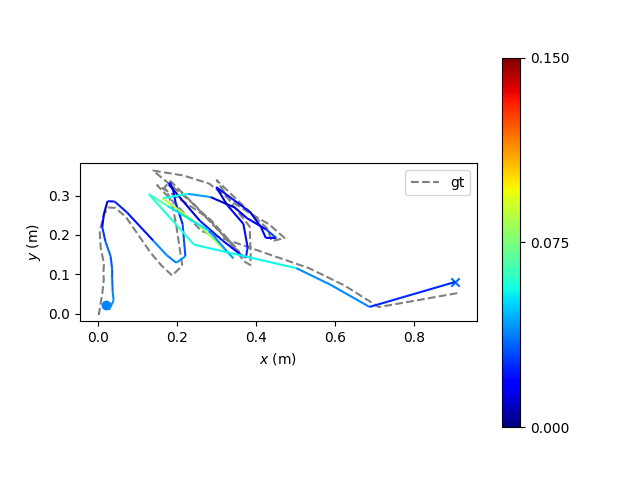} &
    \includegraphics[width=0.22\linewidth]{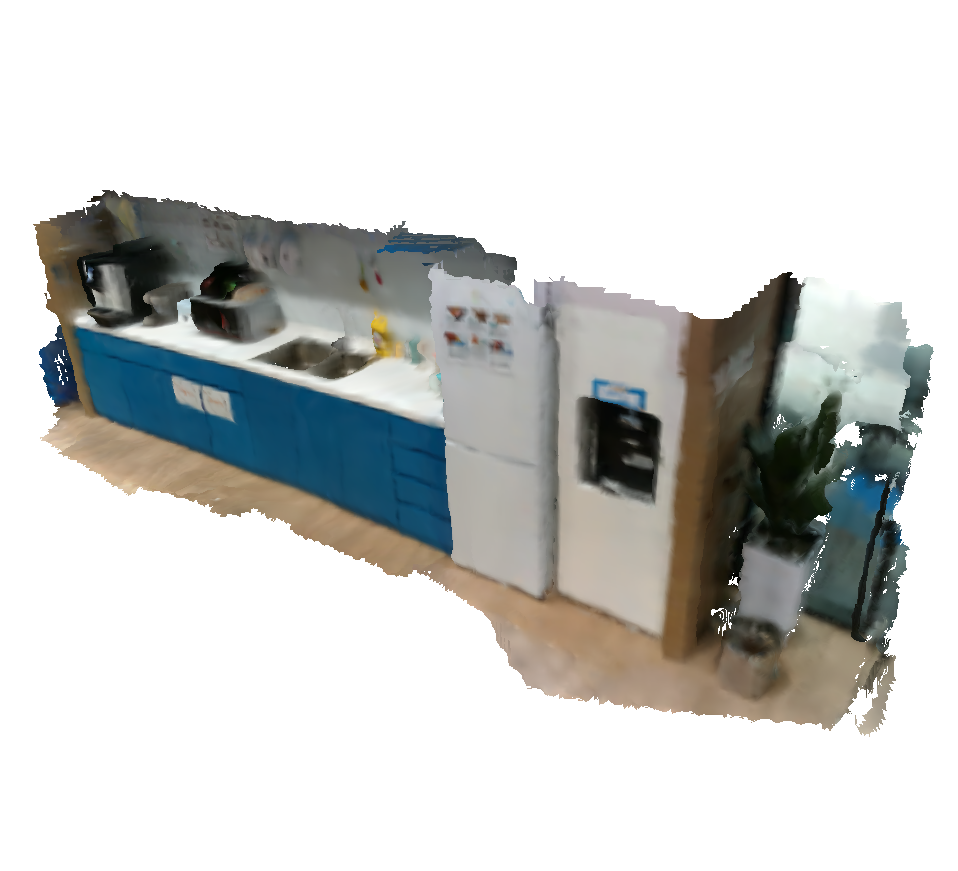} &
    \includegraphics[width=0.22\linewidth]{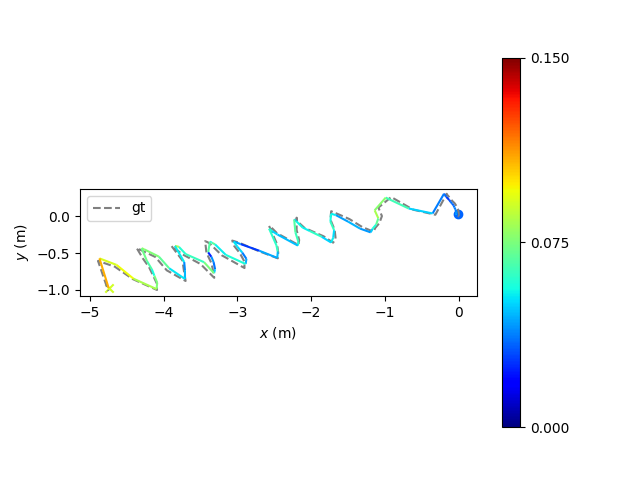} \\
    \multicolumn{2}{c}{\scriptsize{Office}} & \multicolumn{2}{c}{\scriptsize{Living Room}} \\
    \includegraphics[width=0.22\linewidth]{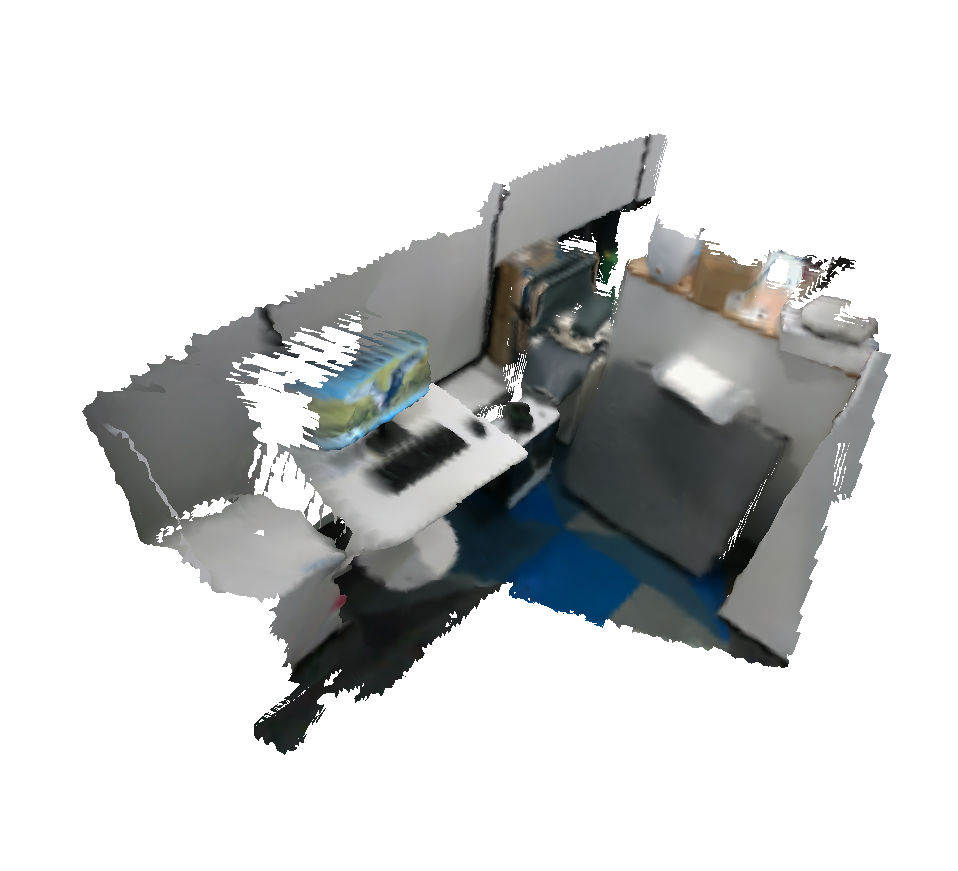} &
    \includegraphics[width=0.22\linewidth]{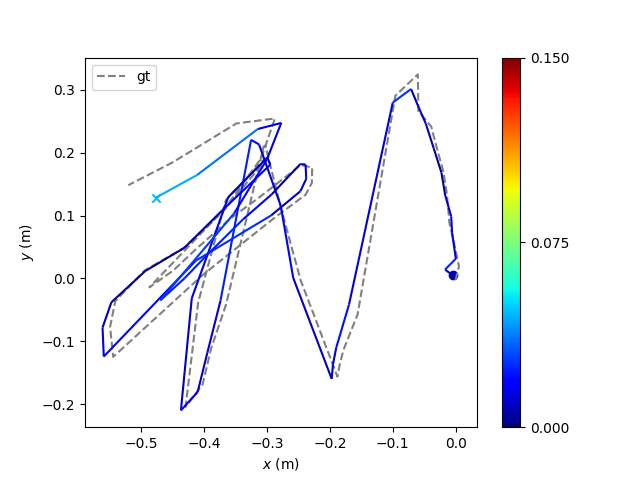} &
    \includegraphics[width=0.22\linewidth]{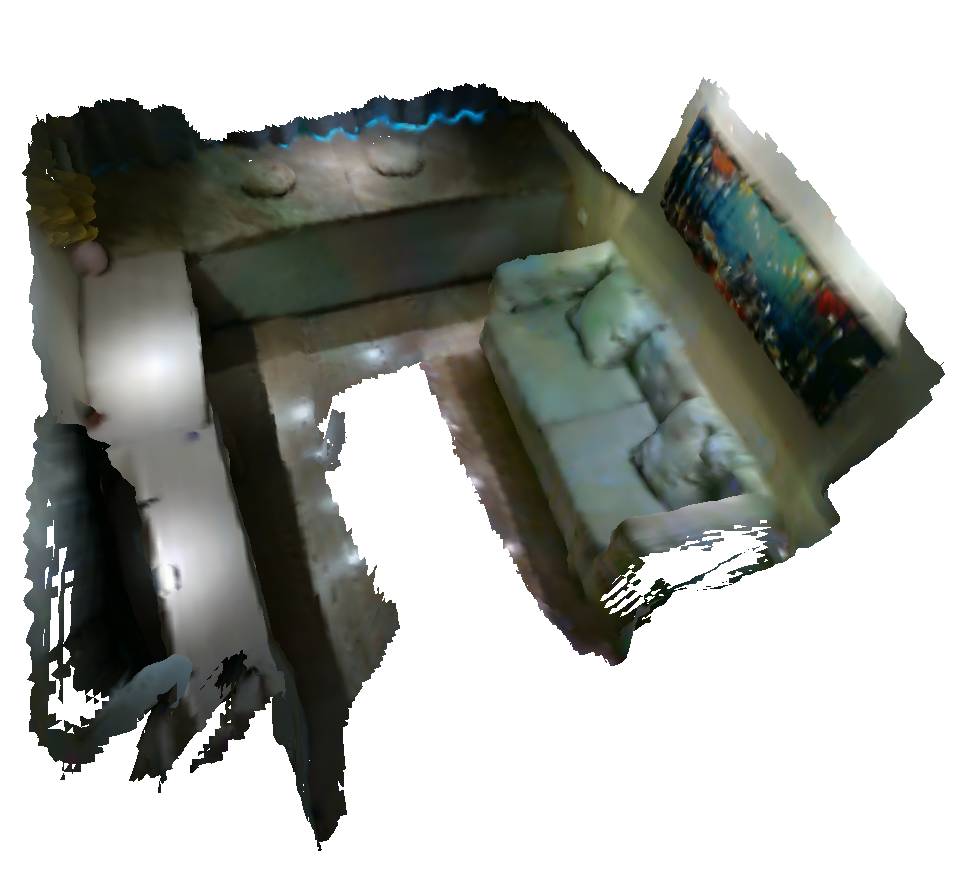} &
    \includegraphics[width=0.22\linewidth]{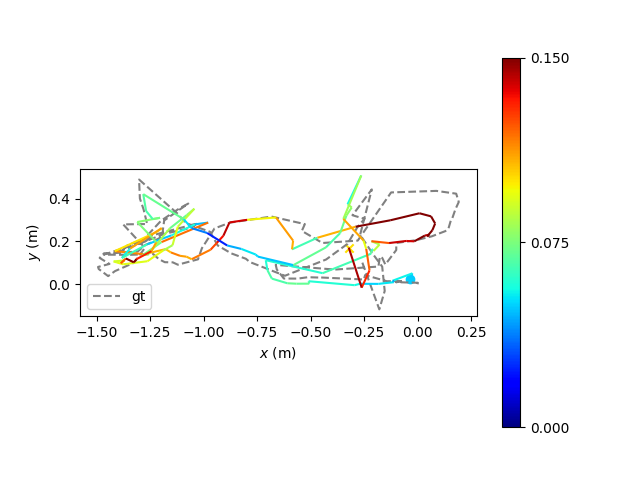} \\
    \multicolumn{2}{c}{\scriptsize{Sofa}} & \multicolumn{2}{c}{\scriptsize{Sofa 2}} \\
    \includegraphics[width=0.22\linewidth]{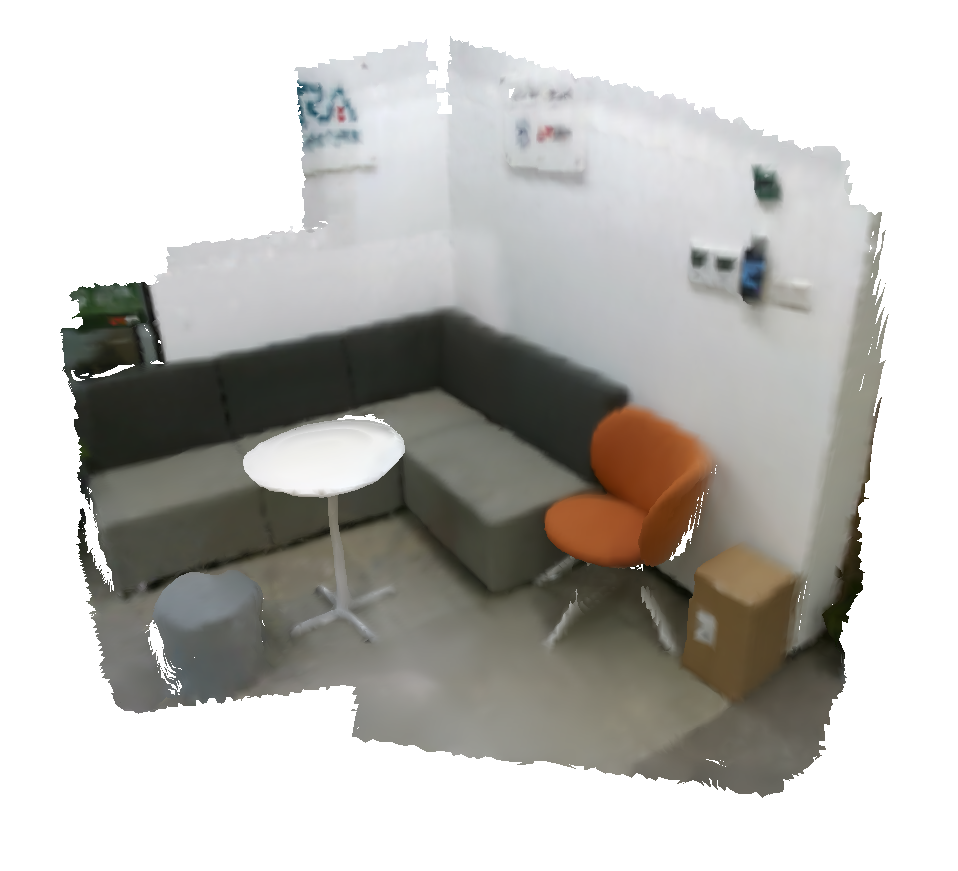} &
    \includegraphics[width=0.22\linewidth]{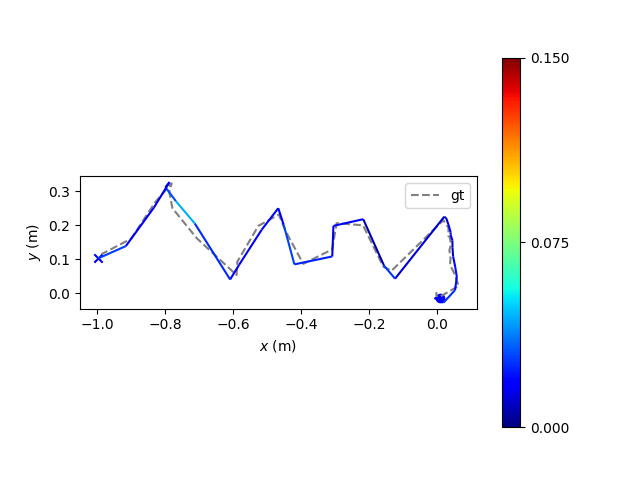} &
    \includegraphics[width=0.22\linewidth]{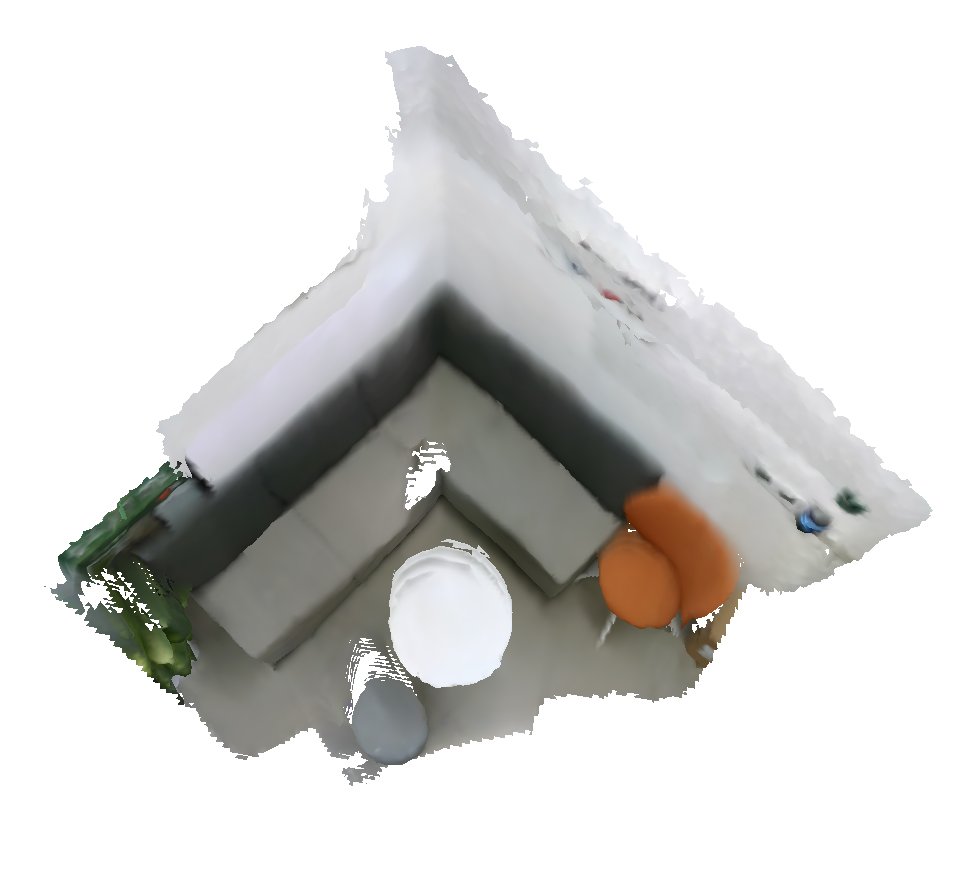} &
    \includegraphics[width=0.22\linewidth]{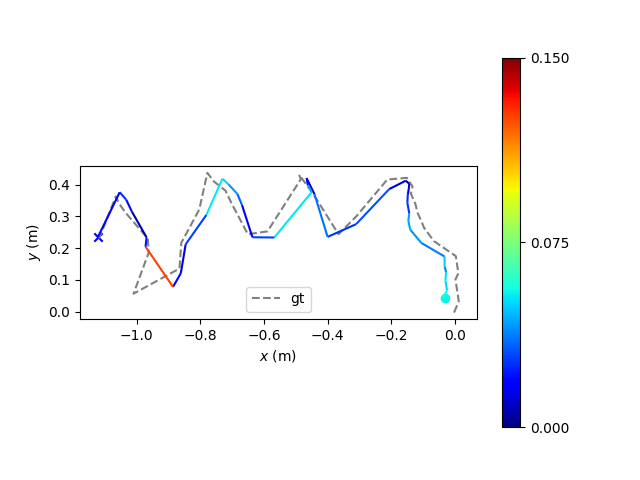} \\
    \end{tabular}
    \caption{\textbf{ZJUL5 Qualitatives.} We provide the trajectory and mesh reconstruction qualitative results on the scenes provided by ZJUL5. \method enables effective mesh reconstructions and accurate tracking.}
    \label{fig:zjul5-supplementary-qualitatives}
\end{figure*}

Figure \ref{fig:zjul5-supplementary-qualitatives} illustrates the reconstructed meshes and predicted trajectories for the scenes included in the ZJUL5 \cite{tof-slam} dataset. Unlike the Replica dataset \cite{replica19arxiv}, which primarily focuses on synthetic environments, the ZJUL5 dataset presents real-world scenarios that introduce a wide range of practical challenges. These include substantial noise and a high density of outliers resulting from sparse Time-of-Flight (ToF) data, as well as environmental difficulties such as poor texture, suboptimal lighting conditions, and limited fields of view.  Despite these hurdles, \method demonstrates remarkable robustness and adaptability, consistently outperforming competing approaches. It is capable of producing high-quality mesh reconstructions and accurately predicting trajectories even under such adverse conditions. These results highlight the strength and versatility of \method in real-world applications, where it effectively addresses the challenges posed by noisy and sparse data while still delivering meaningful and reliable outputs.

\begin{figure*}
\centering
\begin{tikzpicture}
    \pgfplotsset{every tick label/.append style={font=\tiny}}
    \begin{groupplot}[
        group style={
            group size = 1 by 3
        },
        axis lines = left,
        ymajorgrids=true,
        grid style=dashed,
        scaled ticks = false,
        scale only axis = true,
        xtick=data,
        yticklabel style={/pgf/number format/.cd, fixed, fixed zerofill, precision=2},
        xticklabel style={/pgf/number format/.cd, fixed, fixed zerofill, precision=2},
    ]

    \nextgroupplot[
        axis lines = left,
        title={\tiny \textbf{(a) MAE [m]}},
        grid style=dashed,
        width=0.6\linewidth,
        height=0.2\linewidth,
        xtick=\empty,
        extra x ticks={2, 3, 4, 6, 8},
        extra x tick labels = {2, 3, 4, 6, 8},
    ]
    \addplot+[mark=triangle*] table[x=sparsity,y=mae]{supplementary/experiments/temporal_sparsity.txt};

    \nextgroupplot[
        axis lines = left,
        title={\tiny \textbf{(b) F-score [\%]}},
        grid style=dashed,
        scaled ticks = false,
        minor x tick num=1,
        width=0.6\linewidth,
        height=0.2\linewidth,
        scale only axis = true,
        yticklabel style={/pgf/number format/.cd, fixed, fixed zerofill, precision=1},
        xtick=\empty,
        extra x ticks={2, 3, 4, 6, 8},
        extra x tick labels = {2, 3, 4, 6, 8},
    ]
    \addplot+[mark=triangle*] table[x=sparsity,y=fscore]{supplementary/experiments/temporal_sparsity.txt};
    
    \nextgroupplot[
        axis lines = left,
        title={\tiny \textbf{(c) ATE [m]}},
        grid style=dashed,
        scaled ticks = false,
        minor x tick num=1,
        width=0.6\linewidth,
        height=0.2\linewidth,
        scale only axis = true,
        yticklabel style={/pgf/number format/.cd, fixed, fixed zerofill, precision=2},
        xtick=\empty,
        extra x ticks={2, 3, 4, 6, 8},
        extra x tick labels = {2, 3, 4, 6, 8},
    ]
    \addplot+[mark=triangle*] table[x=sparsity,y=ape]{supplementary/experiments/temporal_sparsity.txt};

    \end{groupplot}
\end{tikzpicture}\vspace{-2ex}
\caption{\textbf{Impact of Temporal Sparsity.} The three line plots represent respectively mean absolute error, F-score, and absolute trajectory error as the subsampling ratio of the ToF frames increases. As temporal sparsity grows, a gradual decline in overall performance is observed, reflecting the reduced availability of depth information. Despite this, \method demonstrates resilience, maintaining reasonable performance by effectively leveraging multi-view cues to mitigate the challenges of sparse depth sampling.
}
\label{fig:temporal-sparsity-study}
\end{figure*}
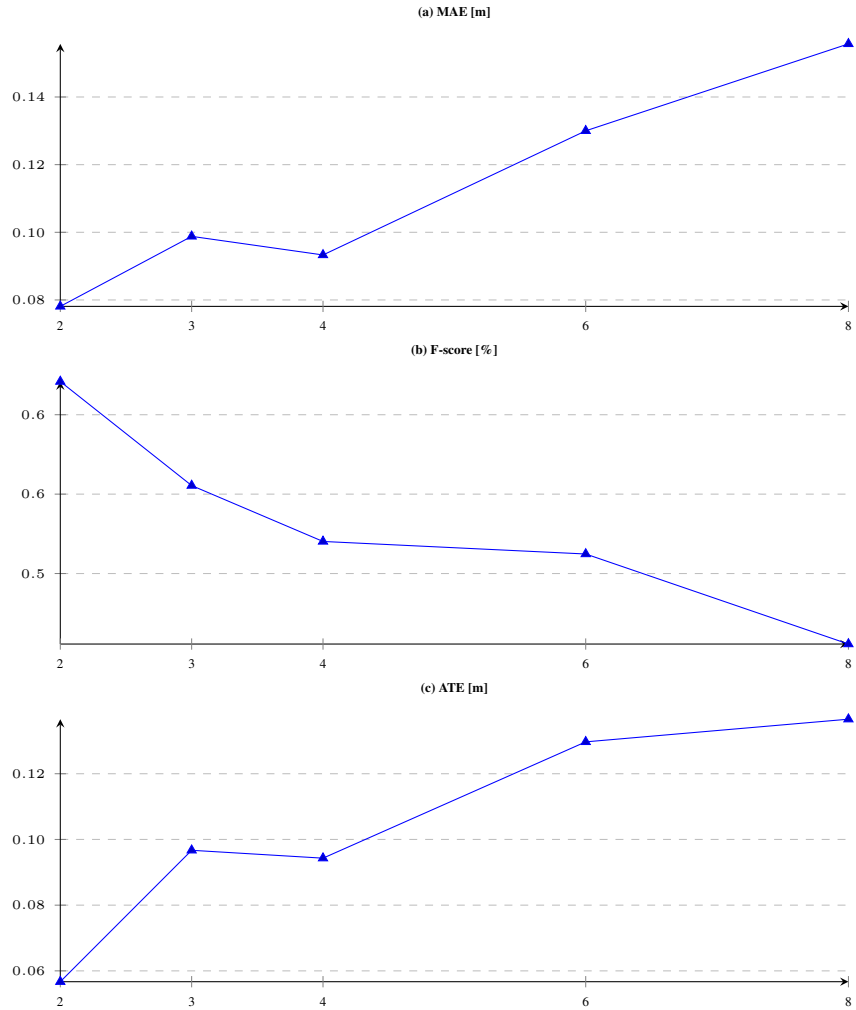

\section{Temporal Sparsity}

Finally, we study \method performance under temporal sparsity, which refers to scenarios where the ToF sensor frame rate is lower than that of the RGB camera. Thus, only a subset of the RGB frames is coupled with sparse depth information. Such a situation is particularly relevant in real use case scenarios where the ToF sensor may operate at a reduced frame rate either due to hardware or power constraints.
To investigate this, we simulate this scenario on the ZJUL5 dataset \cite{tof-slam} by subsampling the ToF frames at ratios of $[2, 3, 4, 6, 8]$, as shown in Figure \ref{fig:temporal-sparsity-study}. \method maintains consistent performance despite the increasing temporal sparsification. The availability of multi-view information enables robustness in the framework, compensating effectively for the lack of sparse depth for a subset of frames.
To ensure that the system has access to the scene scale, we provide ToF depth for every frame in the first 50 frames to provide a reliable starting point. The results highlight the adaptability of \method, demonstrating its ability to operate effectively even when the ToF sensor's frame rate is significantly lower than that of the RGB camera.

\section{Depth on Demand Training Details}

In \method, we perform multi-frame integration adapting the Depth on Demand framework \cite{depth-on-demand} to our specific use case. Specifically, we significantly modify its innermost logic to integrate monocular cues and handle a larger number of frames to overcome the original two-frame configuration. Moreover, we retrain the framework with adjustments designed to optimize its performance in scenarios characterized by extreme input depth sparsity. The architectural improvements made to the framework are detailed in the main paper. To train the model, we utilized the ScanNetV2 dataset \cite{dai2017scannet}, which provides a robust and diverse set of scenes. However, our training procedure diverges from the one described in the original paper \cite{depth-on-demand}. Indeed, at each iteration, we sample a set of source views between 0 and 5 and directly extract sparse depth measurements from the target view instead of performing a reprojection from one of the previous source views. This adjustment aligns better with our use case and eliminates reliance on source-to-target projections for depth data. Additionally, we introduced variability in the density of sparse depth samples, randomly selecting a density within the range $[0\%, \dots, 0.03\%]$. Such sparsification is important since it effectively mimics the extremely sparse depth scenarios encountered in our application, ensuring that the model is robust to real-world conditions of depth sparsity. These enhancements collectively enable \method to achieve high-quality performance, even in environments where data sparsity and noise are significant challenges.

\end{document}